\documentclass{article}
\usepackage{fullpage,appendix}
\usepackage{cite}

\usepackage{microtype}
\usepackage{graphicx}
\usepackage{subfigure}
\usepackage{booktabs} 
\usepackage{multirow}
\usepackage[table,xcdraw]{xcolor}
\usepackage{makecell,enumitem}

\usepackage{hyperref}

\usepackage{amsmath}
\usepackage{amssymb}
\usepackage{mathtools}
\usepackage{amsthm}
\usepackage{algpseudocode}
\usepackage{algorithm}

\usepackage[capitalize,noabbrev]{cleveref}

\newtheorem{theorem}{Theorem}

\newtheorem{definition}{Definition}
\newtheorem{proposition}{Proposition}

\allowdisplaybreaks

\usepackage[textsize=tiny]{todonotes}
\newcommand{\STATE}{\State}

\title{An Interpretable Evaluation of Entropy-based Novelty \\ of Generative Models}

\date{}

\author{
Jingwei~Zhang\thanks{Department of Computer Science and Engineering, The Chinese University of Hong Kong, \url{jwzhang22@ cse.cuhk.edu.hk}}~,
Cheuk Ting Li\thanks{Department of Information Engineering, The Chinese University of Hong Kong, \url{ctli@ie.cuhk.edu.hk}}~,
Farzan~Farnia\thanks{Department of Computer Science and Engineering, The Chinese University of Hong Kong, \url{farnia@cse.cuhk.edu.hk}}
	}     

\begin{document}
\maketitle

\begin{abstract}
The massive developments of generative model frameworks require principled methods for the evaluation of a model's novelty compared to a reference dataset. While the literature has extensively studied the evaluation of the quality, diversity, and generalizability of generative models, the assessment of a model's novelty compared to a reference model has not been adequately explored in the machine learning community. In this work, we focus on the novelty assessment for multi-modal distributions and attempt to address the following \emph{differential clustering} task: Given samples of a generative model $P_\mathcal{G}$ and a reference model $P_\mathrm{ref}$, how can we discover the sample types expressed by $P_\mathcal{G}$ more frequently than in $P_\mathrm{ref}$? We introduce a spectral approach to the differential clustering task and propose the \emph{Kernel-based Entropic Novelty (KEN)} score to quantify the mode-based novelty of $P_\mathcal{G}$ with respect to $P_\mathrm{ref}$. We analyze the KEN score for mixture distributions with well-separable components and develop a kernel-based method to compute the KEN score from empirical data. We support the KEN framework by presenting numerical results on synthetic and real image datasets, indicating the framework's effectiveness in detecting novel modes and comparing generative models. The paper's code is available at: \url{www.github.com/buyeah1109/KEN}.     
\end{abstract}

\section{Introduction}
\label{sec:intro}
Deep generative models including variational autoencoders (VAEs) \cite{kingma2013auto}, generative adversarial networks (GANs) \cite{goodfellow2014generative}, and denoising diffusion models \cite{ho2020denoising} have attained remarkable results in many machine learning problems. The success of these models is primarily due to the great capacity of deep neural networks to express the complex distributions of image, audio, and text data. The impressive qualitative results of deep generative models have inspired several theoretical and empirical studies on their evaluation to reveal the advantages and disadvantages of existing architectures for training generative models.

To compare different generative modeling schemes, several evaluation metrics have been proposed in the literature. The existing evaluation scores can be classified into two categories: 1) distance-based metrics such as Fréchet Inception Distance (FID) \cite{heusel2017gans} and Kernel Inception Distance (KID) \cite{binkowski2018demystifying} measuring the closeness of the distribution of data and generative model, 2) quality, diversity, and generalizability scores such as Inception score \cite{Salimans2016}, Precision/Recall \cite{sajjadi2018assessing,kynkaanniemi2019improved}, and Density/Coverage \cite{naeem2020reliable} assessing the sharpness and variety of the generated data. The mentioned metrics tend to assign higher scores to models closer to the underlying data distribution. While such a property is desired in the evaluation of a learning framework, it may not result in an assessment of a generative model's novelty compared to a baseline generative model or another reference distribution. 

However, the massive developments of generative models highlight the need to assess a model's novelty  
compared to other  
models, because an interpretable comparison between generative models requires the identification of sample types generated by one model more frequently than by the other models. Moreover, prompt-based generative models are often utilized to follow the user's input text prompts to create novel contents, e.g. images of a novel scene or object. If the goal is to maximize the uncommonness of the generated data compared to a reference dataset, a relevant evaluation factor is the model's expressed novelty in comparison to the reference distribution.    

In this work, we focus on the novelty evaluation task in the context of multi-modal distributions which are often present in large-scale image and text datasets due to the different background features of the collected data. In our theoretical analysis, we suppose the test and reference models consist of multiple modes and aim to solve a \emph{differential clustering} task for identifying the novel modes produced by the test model more frequently than by the reference distribution. We propose a \emph{spectral approach} to the differential clustering problem by analyzing the \textit{kernel covariance matrix} of the test and reference distributions, yielding eigenvalues measuring the frequency of differently-expressed modes and eigenvectors revealing the detected modes' sample clusters.  

In the proposed spectral framework, we attempt to compute the eigenspace of the kernel covariance matrices of the test and reference distributions. Assuming the Gaussian kernel with a properly chosen bandwidth, we 
prove that the eigenvalues and eigenvectors of the kernel covariance matrix will approximate the frequency and mean of the modes in a mixture distribution with well-separable components. 
Based on this result, we analyze the eigenspectrum of matrix 
 $\Lambda_{\mathbf{X}|\eta\mathbf{Y}}=C_{\mathbf{X}} -\eta C_{\mathbf{Y}}$, i.e. the difference between the kernel covariance matrices of test $\mathbf{X}$ and reference $\mathbf{Y}$ data multiplied by coefficient $\eta \ge 1$. We demonstrate the application of $\Lambda_{\mathbf{X}|\eta\mathbf{Y}}$'s eigendecomposition to identify the novel modes of test data $\mathbf{X}$ with an $\eta$-times higher frequency than in the reference data $\mathbf{Y}$. As a result, to quantify the mode-based novelty, we propose computing the entropy of the positive eigenvalues of $\Lambda_{\mathbf{X}|\eta\mathbf{Y}}$, which we define to be the \emph{Kernel-based Entropic Novelty (KEN)}  score.

To compute the KEN score under high-dimensional kernel feature maps, e.g. the Gaussian kernel with an infinite-dimensional feature map, we develop a kernel-based method to compute the matrix's eigenvalues and eigenvectors. The proposed algorithm only requires the knowledge of pairwise kernel similarity scores between the observed $\mathbf{X}$ and $\mathbf{Y}$ samples and circumvents the computation challenges under high-dimensional kernel feature maps. Specifically, for a kernel function $k$, we show the $\eta$-differential kernel covariance matrix $\Lambda_{\mathbf{X}|\eta\mathbf{Y}}$ shares the same eigenspectrum with the following matrix $K_{\mathbf{X}|\eta \mathbf{Y}}$ which we call the \emph{$\eta$-differential kernel matrix}:
\begin{equation*}
    K_{\mathbf{X}|\eta \mathbf{Y}}\: =\: \begin{bmatrix} K_{\mathbf{X}\mathbf{X}}\hspace{1.5mm} & \sqrt{\eta}\, K_{\mathbf{X}\mathbf{Y}}\hspace{1.5mm}\vspace{3mm} \\ -\sqrt{\eta}\, K^{^{\Large\top}}_{\mathbf{X}\mathbf{Y}}\hspace{1.5mm} & -\eta\, K_{\mathbf{Y}\mathbf{Y}}\hspace{1.5mm}\vspace{1mm}  \end{bmatrix}, 
\end{equation*}
where $K_{\mathbf{X}\mathbf{X}}$, $K_{\mathbf{Y}\mathbf{Y}}$, and 
$K_{\mathbf{X}\mathbf{Y}}$
denote the kernel matrices for $\mathbf{x}$ samples, $\mathbf{y}$ samples, and the cross kernel matrix between $\mathbf{x}$ and $\mathbf{y}$ samples, respectively. Also, while this matrix-based approach leads to the eigenvalue computation for the non-Hermitian matrix $K_{\mathbf{X}|\eta \mathbf{Y}}$, we show the application of Cholesky decomposition to reduce the problem to the eigendecomposition of a symmetric matrix, which can be handled more efficiently using standard linear algebra programming packages.

Finally, we present the numerical application of our proposed spectral method to several synthetic and real image datasets. For the synthetic experiments, we apply the novelty quantification and detection method to Gaussian mixture models and show the method can successfully count and identify the additional modes in the test distribution compared to the reference mixture model. In our experiments on real datasets, we apply the proposed method to identify the differently expressed sample clusters between standard image datasets. The numerical results suggest the methods' success in detecting the novel concepts present in the datasets. 
Furthermore, we apply the spectral method to detect the modes expressed with different frequencies by state-of-the-art generative modeling frameworks. 
The following is a summary of this work's main contributions:
\begin{itemize}[leftmargin=*]
    \item Proposing a kernel-based spectral method to analyze and quantify mode-based novelty across multi-modal distributions,
    \item Providing theoretical support for the novelty quantification method on mixture distributions with well-separable components,
    \item Developing a kernel method for computing the proxy mode centers and frequencies under high-dimensional kernel feature maps,
    \item Applying the spectral method to detect differently expressed modes between standard generative models. 
\end{itemize}

\section{Related Work}

\textbf{Fidelity and diversity evaluation of generative models}. The evaluation of generative models has been studied in a large body of related works as surveyed in \cite{Borji2022}. The literature has proposed several metrics for the evaluation of the model's distance to the data distribution \cite{heusel2017gans,binkowski2018demystifying}, quality, and diversity \cite{sajjadi2018assessing,kynkaanniemi2019improved,naeem2020reliable,jalali2023information,dan2023vendi}. Except the reference \cite{jalali2023information}, these works do not focus on mixture distributions. Also, our analysis concerns novelty evaluation between two distributions, different from the diversity assessment task addressed by \cite{jalali2023information}.

{Also,  \cite{stein2023exposing,kyn2023} have demonstrated that standard score-based evaluation methods may lead to a biased evaluation due to the choice of Inception-V3 embedding commonly used for image-based generative models. \cite{stein2023exposing} empirically show the less biased evaluation results using DINOv2 embedding. We note the similar importance of the selection of embedding in the results of the spectral KEN approach. We also highlight that the spectral method for evaluating KEN score results in an interpretable evaluation by identifying novel sample clusters between the test and reference distributions, whose relevance can be investigated by the evaluator.} \\
\textbf{Generalization evaluation of generative models}. Several related works aim to measure the generalizability of generative models from training to test data. 
\cite{alaa2022faithful} use the percentage of authenticity to measure the likelihood of generated data copying the training data. \cite{meehan2020non} analyze training data-copying tendency by comparing the average distance to the closest training and test samples. \cite{jiralerspong2023feature} examine overfitting by comparing the likelihoods based on training and test set. We note that the novelty evaluation task considered in our work is different from the generalizability assessment performed in these works, because our definition of mode-based novelty puts more emphasis on out-of-distribution modes not existing in the reference dataset. Also, we note that the reference dataset in out analysis may not be the training set of generative models, resulting in a different task from a training-to-test generalization evaluation. \\
\textbf{Sample rarity and likelihood divergence}. \cite{han2023rarity} empirically show that rare samples are far from the reference data in the feature space. They propose the rarity score as the nearest-neighbor distance to measure the uncommonness of image samples. Also, \cite{jiralerspong2023feature} measure the difference in likelihood of generated distribution to the training and another reference dataset. They propose measuring the likelihood divergence and 
interpret novelty as low memorization of training samples. 
We note that both these evaluations lead to sample-based scores aiming to measure the uncommonness of a single data point. On the other hand, our proposed novelty evaluation is a distribution-based evaluation where we aim to measure the overall mode-based novelty of a model compared to a reference distribution. 


\section{Preliminaries}

\subsection{Novelty Evaluation of Generative Models}

Consider a generator function $G:\mathbb{R}^r\rightarrow \mathbb{R}^d$ mapping an $r$-dimensional latent vector $\mathbf{Z}$ to $G(\mathbf{Z})$ which is aimed to be a real-like sample mimicking the data distribution $P_{\text{\rm data}}$. Here $\mathbf{Z}$ is drawn according to a known distribution, e.g. an isotropic Gaussian $\mathcal{N}(\mathbf{0},\sigma^2 I)$. However, the probability distribution of random vector $G(\mathbf{Z})$ could be challenging to compute for a neural network $G$. The goal in the evaluation of generative model $P_{G(\mathbf{Z})}$ is to quantify and estimate a desired property of its generated samples, e.g. quality and diversity, from $n$ independently generated samples $G(\mathbf{z}_1),\ldots,G(\mathbf{z}_n)$. 

In this work, we focus on the evaluation of novelty in the generated data compared to a reference distribution $Q$ for a random $d$-dimensional vector $\mathbf{Y}$. We assume we have access to $m$ samples in $\{\mathbf{y}_1,\ldots , \mathbf{y}_m\}$  drawn independently from $Q$. Also, for brevity, we denote the generative model $G$'s generated data by $\mathbf{x}_i = G(\mathbf{z}_i)$ for every $1\le i\le n$. Therefore, our aim is to quantify the novelty of generated dataset $\{\mathbf{x}_1,\ldots ,\mathbf{x}_n\}$ compared to reference dataset $\{\mathbf{y}_1,\ldots , \mathbf{y}_m\}$.

In our theoretical analysis, we assume the generated samples follow a multi-modal distribution. We use $P=\sum_{i=1}^k \omega_i P_i$ to represent a $k$-modal mixture distribution where every component $P_i$ has frequency $\omega_i$. Note that $[\omega_1,\ldots, \omega_k]$ represent a probability model on the $k$ modes in $P$ satisfying $\omega_i\ge 0$ for every $i$ and $\sum_{i=1}^k \omega_i = 1$. We assume each component $P_i$ has $\sigma_i^2$-bounded total variance, defined as the trace of $P_i$'s covariance matrix, meaning that given its mean vector $\boldsymbol{\mu}_i$, we have $\mathbb{E}_{\mathbf{X}\sim P_i}\bigl[\bigl\Vert \mathbf{X}-\boldsymbol{\mu}_i\bigr\Vert^2_2\bigr]\le \sigma^2_i$.

\subsection{Kernel Function and Kernel Covariance Matrix}

Consider a kernel function $k:\mathbb{R}^d\times \mathbb{R}^d\rightarrow \mathbb{R}$ mapping every two vectors $\mathbf{x},\mathbf{y}\in\mathbb{R}^d$ to a similarity score $k(\mathbf{x},\mathbf{y})$ satisfying the positive semi-definite (PSD) property, i.e. the kernel matrix $K=\bigl[k(\mathbf{x}_i,\mathbf{x}_j)\bigr]_{n\times n}$ is a symmteric PSD matrix for every selection of input vectors $\mathbf{x}_1,\ldots ,\mathbf{x}_n$. In this paper, we commonly suppose a normalized Gaussian kernel $k_{G(\sigma)}$ with bandwidth parameter $\sigma$ defined as
\begin{equation*}
    k_{G(\sigma)}\bigl(\mathbf{x},\mathbf{y} \bigr)\, := \, \exp\Bigl(\frac{-\Vert \mathbf{x}- \mathbf{y} \Vert^2_2}{2\sigma^2} \Bigr). 
\end{equation*}
 We remark that the PSD property of a kernel function $k$ is equivalent to the existence of a feature map $\phi:\mathbb{R}^d\rightarrow \mathbb{R}^s$ such that for every input vectors $\mathbf{x,y}$ we have $
    k\bigl(\mathbf{x},\mathbf{y} \bigr) = \bigl\langle \phi(\mathbf{x}),\phi(\mathbf{y}) \bigr\rangle
$, where $\langle \cdot ,\cdot \rangle$ denotes the inner product in $\mathbb{R}^s$. Also, we call a kernel function $k$ normalized if for every $\mathbf{x}\in\mathbb{R}^d$ $k(\mathbf{x},\mathbf{x})=1$, e.g. in the defined Gaussian kernel.


Given a distribution $P$ with probability density function $p(\mathbf{x})$ on $\mathbf{X}\in\mathbb{R}^d$, we define the kernel covariance matrix according to kernel $k$ with feature map $\phi$ as\vspace{-1mm}
\begin{align*}
    C_{\mathbf{X}} \, :=&\, \mathbb{E}_{\mathbf{X}\sim P}\bigl[ \phi(\mathbf{X})\phi(\mathbf{X})^\top\bigr] \, =\, \int p(\mathbf{x})\phi(\mathbf{x})\phi(\mathbf{x})^\top \mathrm{d}\mathbf{x}.
\end{align*}
Using the empirical distribution $\widehat{P}_n$ of $n$ samples $\mathbf{x}_1,\ldots ,\mathbf{x}_n$ the kernel covariance matrix will be\vspace{-1mm}
\begin{equation*}
    \widehat{C}_{\mathbf{X}} = \frac{1}{n}\sum_{i=1}^n \phi(\mathbf{x}_i)\phi(\mathbf{x}_i)^\top,
\end{equation*}
which can be written as $\widehat{C}_{\mathbf{X}} = \frac{1}{n}\Phi_{\mathbf{X}}\Phi_{\mathbf{X}}^\top$ that $\Phi_{\mathbf{X}}$ is an $n\times s$ matrix with every $i$-th row being $\phi(\mathbf{x}_i)$.
\begin{proposition}
Using the above definitions, $\widehat{C}_{\mathbf{X}}$ shares the same eigenvalues with the $n\times n$ normalized kernel matrix $\frac{1}{n}\bigl[k(\mathbf{x}_i,\mathbf{x}_j)\bigr]_{n\times n}$ with every $(i,j)$th entry being $\frac{1}{n}k(\mathbf{x}_i,\mathbf{x}_j)$. Therefore, assuming a normalized kernel function, the eigenvalues of $\widehat{C}_{\mathbf{X}}$ are non-negative and sum up to $1$. 
\end{proposition}

\section{A Spectral Approach to Novelty Evaluation for Mixture Models}
In this section, we propose a spectral approach to the novelty evaluation of a generated $\mathbf{X}$ with mixture distribution $P=\sum_{i=1}^k \omega_i P_i$ in comparison to a reference $\mathbf{Y}$ distributed according to mixture model $Q=\sum_{i=1}^t \gamma_i Q_i$. In what follows, we first define and intuitively explain the proposed novelty evaluation score, and later in this section, we will provide a theoretical analysis of the framework in a setting where the mixture models consist of well-separable components. 

To define the proposed novelty score, we focus on the kernel covariance matrices of $\mathbf{X}$ and $\mathbf{Y}$, denoted by $C_{\mathbf{X}}$ and $C_{\mathbf{Y}}$, respectively. We will show in this section that under a Gaussian kernel with proper bandwidth, $C_{\mathbf{X}}$ and $C_{\mathbf{Y}}$ will contain the information of the modes in their eigendecomposition where the eigenvalues can be interpreted as the mode frequencies. Here, given a parameter $\eta\ge 1$, we define the \emph{$\eta$-differential kernel covariance matrix} $ \Lambda_{\mathbf{X}|\eta \mathbf{Y}}$ as follows:
\begin{equation}\label{Eq: difference covariance matrix}
    \Lambda_{\mathbf{X}|\eta \mathbf{Y}}\,  := \, C_{\mathbf{X}} - \eta C_{\mathbf{Y}}. 
\end{equation}

Note that if the components $P_1,\ldots , P_r$ are shared between $P$ and $Q$, they will get canceled in the calculation of $ \Lambda_{\mathbf{X}|\eta \mathbf{Y}}$ and thus will not result in a positive eigenvalue in the eigendecomposition of $\Lambda_{\mathbf{X}|\eta \mathbf{Y}}$ unless their frequency in $P$ is greater than $\eta$-times their frequency in $Q$. This shows how we can, loosely speaking, ``subtract $Q$ from $P$'' by subtracting their kernel covariance matrices in the above definition. Here, the hyperparameter $\eta \ge 1$ controls how much more frequent a mode in $P$ must be compared to the corresponding mode in $Q$ in order to be taken into account.

Therefore, we use the positive eigenvalues of $\Lambda_{\mathbf{X}|\eta \mathbf{Y}}$ to approximate the relative frequencies of the modes of $P$ that are expressed at least $\eta$-times more frequently than in $Q$. This allows us to define the following entropic score to quantify the novelty of $\mathbf{X}$ with respect to $\mathbf{Y}$.
\begin{definition}
 Consider the positive eigenvalues $\lambda_1,\ldots , \lambda_{k'} > 0$ of $\Lambda_{\mathbf{X}|\eta \mathbf{Y}}= C_{\mathbf{X}} - \eta C_{\mathbf{Y}}$ and let $S=\sum_{i=1}^{k'} \lambda_i$. The {Kernel Entropic Novelty (KEN)} score is\vspace{-1mm}
 \begin{equation}\label{Eq: KEN score}
     \text{\rm KEN}_\eta(\mathbf{X}|\mathbf{Y})\: := \: \sum_{i=1}^{k'} \lambda_i \log \frac{S}{\lambda_i}.
 \end{equation}\vspace{-2mm}
\end{definition}\vspace{-2mm}

Intuitively, $\lambda_i/S$ is the relative frequency of the $i$-th novel mode in $P$, and the entropy of the novel mode distribution is $\sum_i (\lambda_i/S) \log (S/\lambda_i)$. The entropy is multiplied by $S$ in the definition of $\text{\rm KEN}_\eta(\mathbf{X}|\mathbf{Y})$ for two reasons. First, the amount of novelty should not only increase with the entropy (or diversity) of the novel modes, but also increase with the total frequency $S$ of those modes. Second, this allows us to interpret $\text{\rm KEN}_\eta(\mathbf{X}|\mathbf{Y})$ as the conditional entropy of the information
of the mode of $\mathbf{X}$ given the dataset of $\mathbf{Y}$.
Later in this section, we will present theoretical results that justify the above informal discussions, and the interpretation of $\text{\rm KEN}_\eta(\mathbf{X}|\mathbf{Y})$ as a conditional entropy.

\subsection{Theoretical Analysis of the Proposed Spectral Novelty Evaluation}

To theoretically analyze the proposed spectral method, we first focus on a single distribution $P$ with well-separable modes and show a relationship between the modes of $P$  and the eigendecomposition of its kernel covariance matrix. We defer the proof of the theoretical results to the Appendix.
\begin{theorem}\label{Thm: Eigendecomposition Single Distribution}
Suppose that every component $P_i$ of a mixture distribution $P=\sum_{i=1}^k \omega_i P_i$ has mean vector $\boldsymbol{\mu}_i$ and bounded total variance  $\mathbb{E}_{\mathbf{X}\sim P_i}\bigl[\Vert \mathbf{X} - \boldsymbol{\mu}_i\Vert^2_2\bigr]\le \sigma_i^2$. Assume that $\omega_1\ge \omega_2\ge \cdots \ge \omega_k$ are sorted in a descending order. Then, the top $k$ eigenvalues $\lambda_1\ge \cdots \ge \lambda_k$ of the kernel covariance matrix $C_{\mathbf{X}}$ according to a Gaussian kernel with bandwidth $\sigma$ will satisfy:\vspace{-1mm}
\begin{align*}
    \sum_{i=1}^k \bigl(\lambda_i - \omega_i \bigr)^2 \; \le&\; 4\sum_{i=1}^k \omega_i \frac{\sigma^2_i}{\sigma^2} +  16\sum_{i=2}^k\sum_{j=1}^{i-1} \omega_i\exp\Bigl(\frac{-\Vert\boldsymbol{\mu}_i - \boldsymbol{\mu}_j \Vert^2_2}{\sigma^2} \Bigr). 
\end{align*}
\end{theorem}
The above theorem shows that if the modes of the mixture distribution are well-separable, meaning that $\min_{1\le i \neq j\le k} \frac{\Vert \boldsymbol{\mu_i} - \boldsymbol{\mu_j} \Vert_2}{\sigma} \gg 1$ while the total variance of the components satisfies $\frac{\sigma_i}{\sigma}\ll 1$, then the eigendecomposition of the Gaussian kernel covariance matrix can reveal the mode frequencies via the principal eigenvalues.

Given the interpretation provided by Theorem~\ref{Thm: Eigendecomposition Single Distribution}, the positive eigenvalues of $\Lambda_{\mathbf{X}|\eta \mathbf{Y}}=C_{\mathbf{X}} - \eta C_{\mathbf{Y}}$ will correspond to the modes of $\mathbf{X}$ that have a frequency at least $\eta$ times higher than the frequency of that mode in $\mathbf{Y}$. Therefore, the positive eigenvalues of $\Lambda_{\mathbf{X}|\eta \mathbf{Y}}$ can be used to quantify the novelty of $\mathbf{X}$ compared to $\mathbf{Y}$. The following theorem formalizes this intuition and shows how $\Lambda_{\mathbf{X}|\eta \mathbf{Y}}$'s positive eigenvalues explain the novelty of $\mathbf{X}$'s modes.
\begin{theorem}\label{Thm: Eigendecomposition Two Distributions}
Consider multi-modal random vectors $\mathbf{X}\sim \sum_{i=1}^k \omega_i P_i$ and $\mathbf{Y}\sim \sum_{i=1}^k \gamma_i Q_i$, where $\omega_{1}-\eta \gamma_{1}\ge \cdots\ge \omega_{k}-\eta \gamma_{k}$. 
Suppose the corresponding mode to every $P_i$ with mean $\boldsymbol{\mu}_i$ is $Q_i$ with mean $\boldsymbol{\mu}'_i = \boldsymbol{\mu}_i + \boldsymbol{\delta}_i$. Then, assuming that for every $i$, both $Q_i$ and $P_i$ have total variance bounded by $\sigma_i^2$, the positive eigenvalues $\lambda_1\ge \ldots \ge \lambda_{k'}>0$ of $\Lambda_{\mathbf{X}|\eta \mathbf{Y}}$ satisfy (letting $\lambda_i = 0$ if $i > k'$)
\begin{align*}
    \sum_{i=1}^{k} \Bigl(\lambda_i - \max\bigl\{\omega_{i}-\eta \gamma_{i} ,0\bigr\}  \Bigr)^2 
    \le\;\; &8\sum_{i=1}^k\Bigl[   \omega_i\frac{ \Vert\boldsymbol{\delta}_i\Vert_2^2}{\sigma^2} +\bigl(\omega_i+\eta^2\gamma_i\bigr)\frac{\sigma_i^2 }{\sigma^2}\Bigr] + \\ &16(1+\eta) \sum_{i=2}^k\sum_{j=1}^{i-1} (\omega_i+\eta\gamma_i)\exp\Bigl(\frac{-\Vert\boldsymbol{\mu}'_i - \boldsymbol{\mu}'_j \Vert^2_2}{\sigma^2} \Bigr).
\end{align*}
\end{theorem}
Based on the above theorem, the principal positive eigenvalues of the defined matrix $\Lambda_{\mathbf{X}|\eta \mathbf{Y}}$ show the extra frequencies of the modes with a more dominant presence in $\mathbf{X}$. As we increase the value of $\eta$, we require a stronger presence of $\mathbf{X}$'s modes to count them as a novel mode. In the limit case where $\eta \rightarrow +\infty$, we require a complete absence of an $\mathbf{X}$'s mode in $\mathbf{Y}$ to call it novel. 

\begin{figure*}
    \centering  \includegraphics[width=\textwidth]{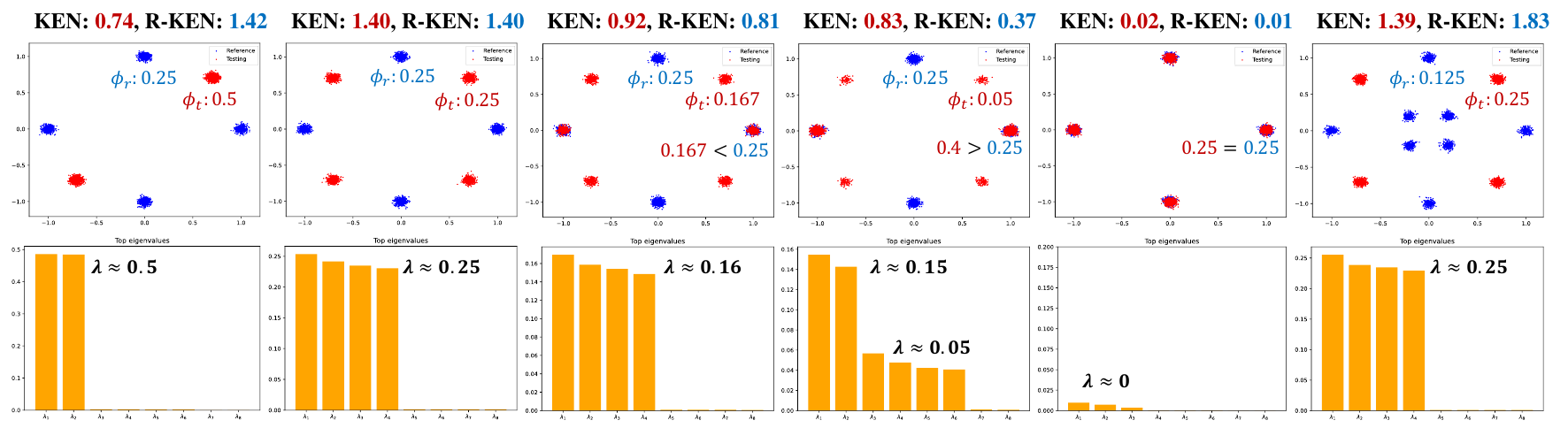}
    \caption{Experimental results on synthetic Gaussian mixture distributions including KEN and R-KEN (Reversed-KEN) scores, and principal eigenvalues of the differential kernel covariance matrix $\Lambda_{\mathbf{X}|\eta \mathbf{Y}}$. \textbf{Top row: } Reference (in blue) and test (in red) samples with $\phi_t,\, \phi_r$ denoting the test and reference modes' frequency. \textbf{Bottom row: } Positive eigenvalues of $\Lambda_{\mathbf{X}|\eta \mathbf{Y}}$.}
    \label{fig:gm}
\end{figure*}

Finally, note that $\mathrm{KEN}_{\eta}(\mathbf{X}|\mathbf{Y})$ can be interpreted as the conditional entropy of the information
of the mode of $\mathbf{X}$ given the dataset of $\mathbf{Y}$. More precisely, if $\eta=1$
this score is the conditional entropy $H(X_{\mathrm{mode}}|Y_{\mathrm{adv}})$, where $X_{\mathrm{mode}} \in \{1,\ldots,k\}$
is the mode cluster variable of $\mathbf{X}$ (the index of the mode $\mathbf{X}$ belongs to), 
and $Y_{\mathrm{adv}}$ represents
the knowledge of an adversary who knows the dataset of $\mathbf{Y}$ and wants
to predict $X_{\mathrm{mode}}$. If the random sample $\mathbf{X}$ is also found
in the dataset of $\mathbf{Y}$, then the adversary will know the
mode of $\mathbf{X}$ and predicts $Y_{\mathrm{adv}}=X_{\mathrm{mode}}$
accurately; otherwise the adversary knows nothing about $X_{\mathrm{mode}}$,
and outputs $Y_{\mathrm{adv}}=e$ as an erasure symbol denoting
the lack of information.

Under the setting in Theorem~\ref{Thm: Eigendecomposition Two Distributions} for $\eta = 1$, $\lambda_{i}=\max\{\omega_{i}-\gamma_{i},0\}$,
take
$\mathbb{P}(Y_{\mathrm{adv}}\! =\! i|X_{\mathrm{mode}}\! =\! i)=\min\{\gamma_{i}/\omega_{i},1\}$ (otherwise $Y_{\mathrm{adv}}=e$) since among the samples of $\mathbf{X}$ with mode $i$, at most a portion $\gamma_{i}/\omega_{i}$ are also samples of $\mathbf{Y}$ (if sizes of the two datasets are equal). We have 
$\mathbb{P}(Y_{\mathrm{adv}}=e)=\sum_{i=1}^{t}\omega_{i}\max\{1-\gamma_{i}/\omega_{i},\,0\}
=S$.
Since $H(X_{\mathrm{mode}}|Y_{\mathrm{adv}}=i)=0$, we have 
\begin{align*}
H(X_{\mathrm{mode}}|Y_{\mathrm{adv}}) & =\mathbb{P}(Y_{\mathrm{adv}}=e)H(X_{\mathrm{mode}}|Y_{\mathrm{adv}}=e)\\
 & =S\sum_{i=1}^{t}\frac{\lambda_{i}}{S}\log\frac{S}{\lambda_{i}} \; =\; \mathrm{KEN}_{1}(\mathbf{X}|\mathbf{Y}).
\end{align*}
For a general $\eta$, $\mathrm{KEN}_{\eta}(\mathbf{X}|\mathbf{Y})$ can be interpreted
as $H(X_{\mathrm{mode}}|Y_{\mathrm{adv}})$ if each sample of $\mathbf{Y}$
allows the adversary to learn $\eta$ samples of $\mathbf{X}$ belonging to
the same mode as $\mathbf{Y}$,
resulting in $\mathbb{P}(Y_{\mathrm{adv}}=i\,|\,X_{\mathrm{mode}}=i)=\min\{\eta\gamma_{i}/\omega_{i},\,1\}$. The next section shows how we can use the kernel trick to compute the KEN score from the pairwise similarity scores between the observed test and reference samples.

\begin{figure*}[t]
    \centering    
    \includegraphics[width=\textwidth]{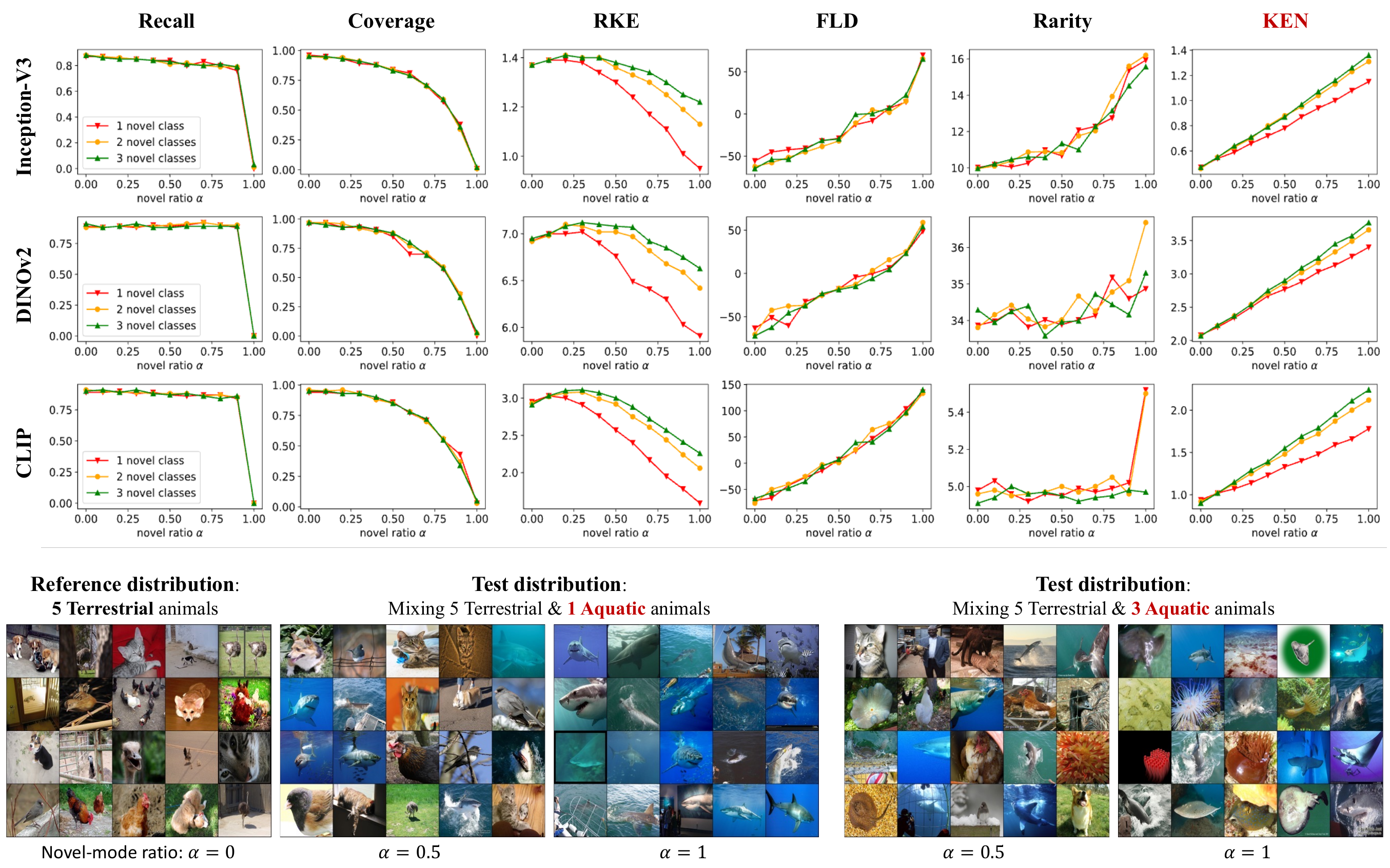}
    \caption{\textbf{Top 3 rows: }Trends of baseline and KEN scores in evaluating novel yet less-diverse distributions with Inception-V3, DINOv2 and CLIP embeddings. \textbf{Bottom: }ImageNet-1K Samples from reference and test distributions. Reference modes: 5 terrestrial animals. Novel modes: 1-3 aquatic lives. $\alpha$ is the ratio of novel modes in testing distribution. $\alpha=0, 1$ represents pure reference and novel distributions, respectively. 
    }
    \label{fig:distinguish}
\end{figure*}

\section{Computation of the KEN Novelty Score}
Since the KEN score is characterized using the difference $C_{\mathbf{X}}- \eta C_{\mathbf{Y}}$ of kernel covariance matrices, the computation of this score will be challenging in the kernel feature space under a high-dimensional kernel feature map. Specifically, the kernel feature map for a Gaussian kernel is infinitely high-dimensional. Our next theorem reduces the eigendecomposition task to the differential kernel matrix based on only the kernel similarity scores of empirical samples. 
\begin{theorem}\label{Thm: Kernel Computation}
Suppose we observed empirical samples $\mathbf{x}_1,\ldots ,\mathbf{x}_n$ from test distribution $P$ and $\mathbf{y}_1,\ldots ,\mathbf{y}_m$ from reference distribution $Q$. Then, the difference of empirical kernel covariance matrices $\widehat{\Lambda}_{\mathbf{X}|\eta \mathbf{Y}} = \widehat{C}_{\mathbf{X}} - \eta \widehat{C}_{\mathbf{Y}}$ shares the same positive eigenvalues with the following matrix, which we call the \emph{$\eta$-differential kernel matrix}:
\begin{equation}\label{Eq: differential kernel matrix}
    K_{\mathbf{X}|\eta \mathbf{Y}}\: :=\: \begin{bmatrix} K_{\mathbf{X}\mathbf{X}}\hspace{1.5mm} & \sqrt{\eta}\, K_{\mathbf{X}\mathbf{Y}}\hspace{1.5mm}\vspace{3mm} \\ -\sqrt{\eta}\, K^\top_{\mathbf{X}\mathbf{Y}}\hspace{1.5mm} & -\eta\, K_{\mathbf{Y}\mathbf{Y}}\hspace{1.5mm}\vspace{1mm}  \end{bmatrix} 
\end{equation}
In the above, $K_{\mathbf{X}\mathbf{X}}=\frac{1}{n}\bigl[k(\mathbf{x}_i,\mathbf{x}_j)\bigr]_{n\times n}$ and $K_{\mathbf{Y}\mathbf{Y}}=\frac{1}{m}\bigl[k(\mathbf{y}_i,\mathbf{y}_j)\bigr]_{m\times m}$ are the kernel similarity matrices for observed $\mathbf{X}$ and $\mathbf{Y}$, samples respectively, and $K_{\mathbf{X}\mathbf{Y}}=\frac{1}{\sqrt{nm}}\bigl[k(\mathbf{x}_i,\mathbf{y}_j)\bigr]_{n\times m}$ is the $n\times m$ cross-kernel matrix between observed ${\mathbf{X},\mathbf{Y}}$ samples. 
\end{theorem}

\begin{algorithm}[t]
    \caption{Computation of KEN \& novel mode centers} 
    \label{algo:KEN}
    \begin{algorithmic}[1]
    \STATE \textbf{Input:} Sample sets $\{\mathbf{x}_1,\ldots,\mathbf{x}_n\}$ and $\{\mathbf{y}_1,\ldots,\mathbf{y}_m\}$, parameter $\eta>0$, Gaussian kernel bandwidth $\sigma$\vspace{1mm}
        \STATE Compute matrices $K_{\mathbf{X}\mathbf{X}}=\frac{1}{n}[k(\mathbf{x}_i,\mathbf{x}_j)]_{n\times n}$, $K_{\mathbf{Y}\mathbf{Y}}=\frac{1}{m}[k(\mathbf{y}_i,\mathbf{y}_j)]_{m\times m}$, $K_{\mathbf{X}\mathbf{Y}}=\frac{1}{\sqrt{mn}}[k(\mathbf{x}_i,\mathbf{y}_j)]_{n\times m}$\vspace{1mm} 
        \STATE Apply Cholesky decomposition to compute upper-triangular $V\in\mathbb{R}^{\tiny(n+m)\times (n+m)}$ such that 
        \begin{equation*}
    V^\top V\, =\, \begin{bmatrix} K_{\mathbf{X}\mathbf{X}} & \sqrt{\eta}\, K_{\mathbf{X}\mathbf{Y}}\vspace{1mm} \\ \sqrt{\eta}\, K^\top_{\mathbf{X}\mathbf{Y}} & \eta\, K_{\mathbf{Y}\mathbf{Y}}\end{bmatrix} 
\end{equation*}
\STATE Compute $\Gamma = V\mathrm{diag}([\underbrace{+1,\ldots , +1}_{n \; \text{\rm times}},\underbrace{-1,\ldots , -1}_{m \; \text{\rm times}}])V^\top$\vspace{1mm}
        \STATE Perform eigendecomposition to get $\Gamma = U^\top \mathrm{diag}(\boldsymbol{\lambda})U $\vspace{1mm} 
    \STATE {Find the positive eigenvalues $\lambda_1 \ge \cdots \ge \lambda_{k'} > 0$ and the corresponding eigenvectors ${\mathbf{u}}_1,\ldots ,{\mathbf{u}}_{k'}$}\vspace{1mm}
    \STATE {Set $\mathbf{u}_i \leftarrow \mathrm{sgn}(\sum_{j=1}^n u_{i,j}) \cdot {\mathbf{u}}_i$ for $i=1,\ldots,k'$}\vspace{1mm}
    \STATE \textbf{Output:} $\text{\rm KEN-score}=\sum_{i=1}^{k'} \lambda_i \log \frac{\sum_{j=1}^{k'} \lambda_j}{\lambda_i}$ and eigenvectors $\mathbf{u}_1,\ldots ,\mathbf{u}_{k'}$.
    \end{algorithmic}
\end{algorithm}

Theorem~\ref{Thm: Kernel Computation} simplifies the eigendecomposition of the matrix $\widehat{\Lambda}_{\mathbf{X}|\eta \mathbf{Y}}$ to the $(n+m)\times (n+m)$ kernel-based matrix $K_{\mathbf{X}|\eta \mathbf{Y}}$. We remark that each eigenvector $\mathbf{v}_i\in\mathbb{R}^{m+n}$ of this matrix contains the expected inner-product of empirical $\mathbf{X}$ and $\mathbf{Y}$ data with the $i$th detected mode, which can be utilized to rank the samples based on their likelihood of belonging to the $i$th identified novel cluster. Due to the sign ambiguity for each eigendirection $\mathbf{v}_i$, we multiply computed eigenvector $\mathbf{v}_i$ by the sign of sum of its first $n$ entries, $\mathrm{sgn}\bigl(\sum_{j=1}^n v_{i,j}\bigr)$, to prefer a positive score for test $\mathbf{x}_1,\ldots,\mathbf{x}_n$ samples. 

While the discussed eigendecomposition can be addressed via $O\bigl((n+m)^3\bigr)$ computations, $K_{\mathbf{X}|\eta \mathbf{Y}}$ is a non-Hermitian matrix, for which standard Hermitian matrix-based algorithms do not apply. In the following theorem, we apply Cholesky decomposition to reduce the task to an eigenvalue computation for a symmetric matrix.

\begin{theorem}\label{Thm: Cholesky}
    In the setting of Theorem \ref{Thm: Kernel Computation}, define the following joint kernel matrix:\vspace{-2mm}
    \begin{equation*}
    K_{\mathbf{X},\eta \mathbf{Y}}\: :=\: \begin{bmatrix} K_{\mathbf{X}\mathbf{X}}\hspace{1.5mm} & \sqrt{\eta}\, K_{\mathbf{X}\mathbf{Y}}\hspace{1.5mm}\vspace{1mm} \\ \sqrt{\eta}\, K^\top_{\mathbf{X}\mathbf{Y}}\hspace{1.5mm} & \eta\, K_{\mathbf{Y}\mathbf{Y}}\hspace{1.5mm}\vspace{1mm}  \end{bmatrix}.
\end{equation*}
Consider the Cholesky decomposition of the above PSD matrix satisfying $K_{\mathbf{X},\eta \mathbf{Y}} = V^\top V$ for upper-triangular matrix $V\in\mathbb{R}^{(n+m)\times (n+m)}$. Then, $\widehat{\Lambda}_{\mathbf{X}|\eta \mathbf{Y}}$ shares the same non-zero eigenvalues with the symmetric matrix $V D V^\top$ where $D$ is a $(n+m)\times (n+m)$ diagonal matrix with diagonal entries in $\bigl[\underbrace{+1,\ldots , +1}_{n \; \text{\rm times}},\underbrace{-1,\ldots , -1}_{m \; \text{\rm times}} \bigr]$.
\end{theorem}
Based on the above theorem, we propose Algorithm~\ref{algo:KEN} to compute the KEN score and find the eigendirections corresponding to the detected novel modes. We note that the eigendecomposition task in the algorithm reduces to the spectral decomposition of a symmetric matrix that can be handled more efficiently than the eigenvalue computation for a general non-symmetric matrix. 

Finally, note that each computed eigenvector $\mathbf{u}_i\in\mathbb{R}^{n+m}$ in Algorithm~\ref{algo:KEN} corresponds to the function $\widetilde{u}_i:\mathbb{R}^d\rightarrow\mathbb{R}$,
\begin{equation*}
    \widetilde{u}_i(\mathbf{x}) = \sum_{j=1}^n u_{i,j}k(\mathbf{x}_j,\mathbf{x}) + \sum_{s=1}^m u_{i,s+n}k(\mathbf{y}_s,\mathbf{x}),
\end{equation*}
 where $u_{i,j}$ stands for the $j$-th entry of $\mathbf{u}_i$. The above function's output $\widetilde{u}_i(\mathbf{x})$ can be viewed as the data point $\mathbf{x}$'s score of belonging to the identified $i$-th cluster.   Therefore, we include Step~7 in Algorithm~\ref{algo:KEN}, which multiplies the computed eigenvector $\mathbf{u}_i$ with $+1$ or $-1$, to prefer a non-negative score for test data $\mathbf{x}_1,\ldots ,\mathbf{x}_n$ in the novelty evaluation task.

\section{Numerical Results}

\subsection{Experimental Setup}

\begin{figure*}
    \centering    \includegraphics[width=0.99\textwidth]{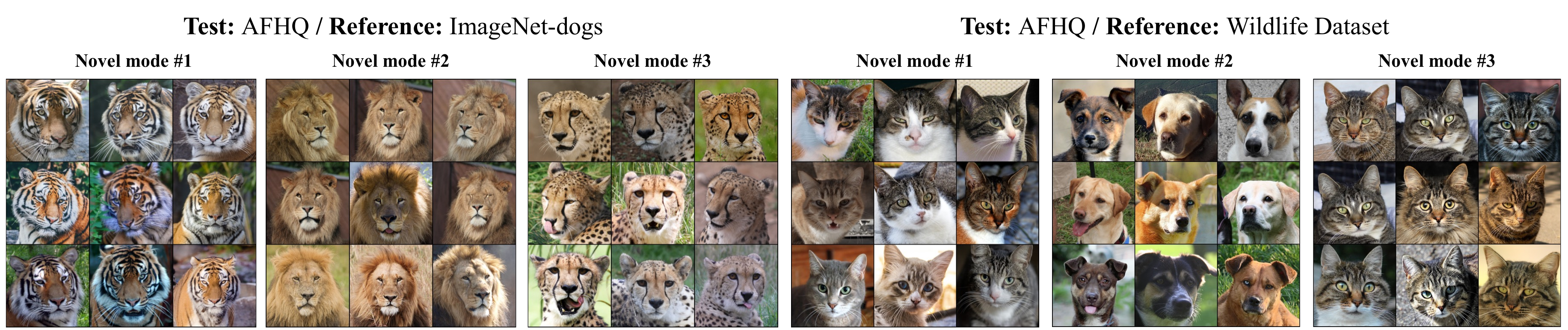}
    
    \caption{Identified top-3 novel modes between image datasets: (Left-half) AFHQ w.r.t. ImageNet-dogs, (Right-half) AFHQ w.r.t. Wildlife. Inception-V3 embedding is used. Shown samples are the test data with the maximum entry values on the top three principal eigenvectors of the differential kernel matrix $K_{\mathbf{X}\vert \eta \mathbf{Y}}$ defined in \eqref{Eq: differential kernel matrix}.}
    \label{fig:visual_realmode}  
\end{figure*}

\textbf{Datasets.} We performed experiments on the following image datasets: 1) CIFAR-10 \cite{krizhevsky2009learning} with 60k images of 10 classes, 2) ImageNet-1K \cite{deng2009imagenet} with 1.4 million images of 1000 classes, containing 20k dog images from 120 different dog breeds, 3) CelebA \cite{liu2015deep} with 200k face images of celebrities,  4) FFHQ \cite{karras2019style} with 70k human-face images, 5) AFHQ \cite{choi2020stargan} with 15k animal-face images of dogs, cats, and wildlife. The AFHQ-dog subset has 5k images from 8 dog breeds. 6) Wildlife dataset \cite{wildlife_dataset} with 2k wild animal images.\\
\textbf{Pre-trained generative models and neural nets for feature extraction:} We used the following embeddings in our experiments: 1) pre-trained Inception-V3 \cite{szegedy2016rethinking} which is the standard in FID and IS scores. 2) DINOv2 \cite{oquab2023dinov2} suggested by \cite{stein2023exposing} to reduce the biases in ImageNet-based Inception-V3 embedding, 3) CLIP \cite{radford2021learning} suggested by \cite{kyn2023} to lessen the inductive biases of Inception-V3 embedding. For a fair comparison between the tested image-based generative models, we downloaded the pre-trained models from the StudioGAN \cite{kang2023StudioGANpami} and \cite{stein2023exposing}'s GitHub repositories. \\ 
\textbf{Bandwidth parameter $\sigma$ and sample size:} Similar to \cite{jalali2023information}, we chose the kernel bandwidth to be the smallest $\sigma$ satisfying variance $<0.01$. In our experiments, we observed $\sigma \in [10, 15]$ could satisfy this requirement for all the tested image data with the Inception-V3 embedding. 
In the case of synthetic Gaussian mixtures, we used $\sigma=0.5$. In our experiments, we used $m,n=5000$ sample size for the test and reference data.

\subsection{Numerical Results on Synthetic Gaussian Mixtures}

First, we tested the proposed methodology on Gaussian mixture models (GMMs) as shown in Figure \ref{fig:gm}. The experiments use the standard setting of 2-dimensional Gaussian mixtures in \cite{gulrajani2017wgan-gp}. We show the samples from the reference distribution (in blue) with a 4-component GMM where the components are centered at [0, 1], [1, 0], [0, -1], [-1, 0]. The generated data in the test distribution (in red) follow a Gaussian mixture in all the experiments, where we center the novel modes (unexpressed in the reference) at [$\pm 0.7$, $\pm 0.7$] and center the shared modes at the same component-means of the reference distribution. In the experiments, we chose parameter $\eta = 1$ for the KEN evaluation.

Based on the KEN scores and eigenspectrum of the $\eta$-differential kernel matrix reported in Figure \ref{fig:gm}, our proposed spectral method successfully identifies the novel modes, and the KEN scores correlate with the novel modes' number and frequencies. We highlight the following trends in the evaluated KEN scores:

\noindent \textbf{1. More novel modes result in a greater KEN score.} The first two columns of Figure \ref{fig:gm} illustrate that adding two novel modes to the test distribution increases the KEN score from 0.74 to 1.40. The bar plots of the differential kernel matrix's eigenvalues also show two extra principal eigenvalues approximating the frequencies of novel modes. \\
\textbf{2. Transferring weight from novel to common modes decreases KEN score.} Columns 3-5 in Figure~\ref{fig:gm} show the effects of overlapping modes on KEN score. In Column 3, the test distribution has six components with uniform frequencies of $1/6$, of which two modes are centered at the same points as the reference modes with frequency $1/4$. The KEN score decreased from 1.40 to 0.92, and we observed only four principal eigenvalues in the differential kernel matrix. Also, when we increased the frequencies of the common modes from 0.25 to 0.4, as shown in Column 4, we could observe 6 outstanding eigenvalues, from which two of them approximate the difference of common mode frequencies. Moreover, under two identical distributions, the KEN score was nearly 0. \\
\textbf{3. KEN does not behave symmetrically between the reference and test distributions.} In our experiments, we also measured the KEN score of the reference distribution with respect to the test distribution, which we call \emph{Reverse KEN (R-KEN)}. We observed that the KEN and R-KEN scores could behave differently, and the roles of test and reference distributions were different. Regarding KEN and R-KEN's mismatch, when we included four extra \emph{reference} modes in the last column, the KEN score did not considerably change (1.40 vs. 1.39) while the R-KEN value jumped from 1.40 to 1.83.

\begin{figure*}
    \centering
    \includegraphics[width=\textwidth]{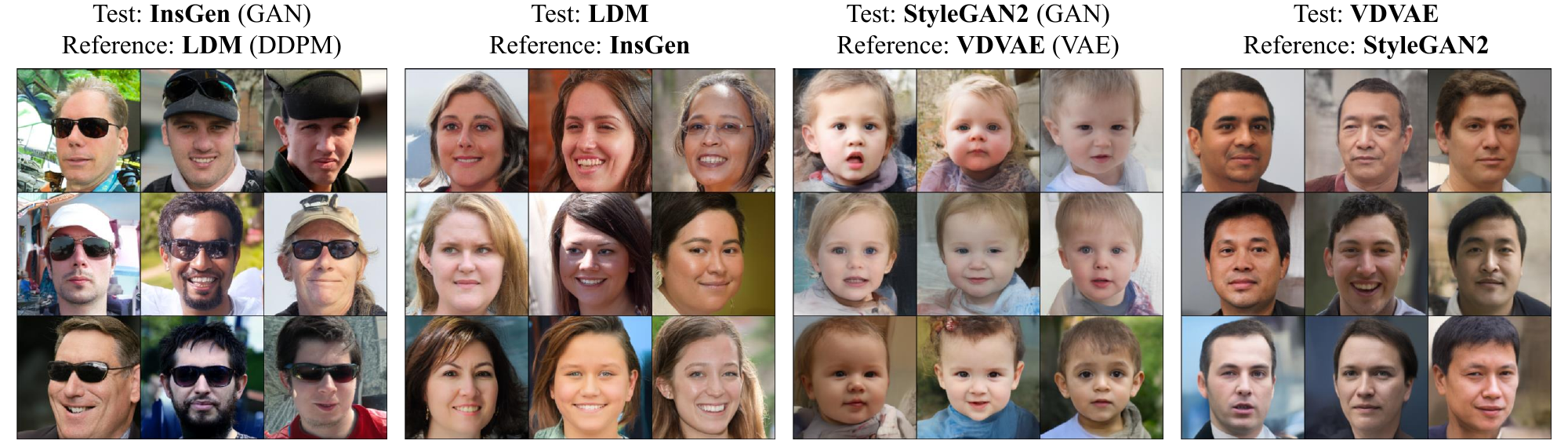}
    
    \caption{Identified top novel modes between FFHQ-trained generative models. Inception-V3 embedding is used.}
    \label{fig:visual_genmode}
\end{figure*}

\subsection{Novelty vs. Diversity Evaluation via KEN and Baseline Metrics}

The novelty and diversity evaluation criteria may not align under certain conditions. To test our proposed method and the existing scores' capability to capture novelty under less diversity, we designed experimental settings where the test distribution possessed less diversity while containing novel modes compared to the reference distribution. The baseline diversity-based scores we attempted in the experiment were improved Recall \cite{Kynk2019}, Coverage \cite{naeem2020reliable}, and RKE \cite{jalali2023information}. We also evaluated the sample divergence: FLD \cite{jiralerspong2023feature} and sample Rarity \cite{han2023rarity}, which are proposed to assess sample-based novelty. We extract images from 5 terrestrial animal classes in ImageNet-1K to form the reference distribution $P_r$ and images from 1-3 aquatic life classes to form the novel distribution $P_n$. To simulate different novelty ratios, we mixed the two distributions with a ratio parameter $0\le \alpha\le 1$ to form the test distribution as $P_t = \alpha P_n + (1-\alpha)P_r$.

As shown in Figure \ref{fig:distinguish}, all the diversity-based baseline metrics decreased under a larger $\alpha$, i.e., as the test distribution $P_t$ becomes closer to the novel distribution $P_n$. The observation can be interpreted as the diversity and novelty levels change in opposite directions in this experiment. The proposed KEN score and baseline FLD and Rarity scores could capture the higher novelty and increased with $\alpha$. On the other hand, when we increased the number of novel modes from 1 to 3 aquatic animals, the sample-based FLD and Rarity scores did not change significantly, while our proposed KEN score could capture the extra novel modes and grew with the number of novel modes. This experiment shows the \emph{distribution-based} novelty evaluation by the KEN score vs. the \emph{sample-based} novelty evaluation by FLD and Rarity.

\begin{table*}
    \caption{FFHQ-trained generative models' pairwise KEN score. Inception-V3 embedding is used.}
    \label{tab:benchmark_novelty}
    \centering
    \scalebox{1}{
    \begin{tabular}{lcccccr}
    \toprule
    \multirow{2}{*}{\makecell{Generative Models \\ (Test Models)}} & \multicolumn{5}{c}{Reference Models}                \\
    \cmidrule(){2-6} 
                                      & InsGen  & LDM  & StyleGAN2 & VDVAE & StyleGAN-XL & Avg. KEN \\
    \midrule
    InsGen \cite{yang2021insgen}                           & -      & 1.26 & 1.18      & 1.87  & 1.17        & 1.37     \\
    LDM \cite{rombach2021highresolution}                              & 1.09   & -    & 1.14      & 1.59  & 1.08        & 1.23     \\
    StyleGAN2 \cite{karras2019style}                        & 1.12   & 1.26 & -         & 1.76  & 1.18        & 1.33     \\
    VDVAE \cite{child2020very}                            & 0.96   & 0.91 & 0.94      & -     & 0.95        & 0.94     \\
    StyleGAN-XL  \cite{sauer2022stylegan}                      & 1.16   & 1.24 & 1.19      & 1.83  & -           & 1.36      \\
    
    \bottomrule
    \end{tabular}}
\end{table*}

\subsection{Numerical Results on Real/Generated Image Data}

We evaluated the KEN score and visualized identified novel modes for the real image dataset and sample sets generated by widely-used generative models.
For the identification of samples belonging to the detected novel modes, we followed Algorithm~\ref{algo:KEN} to obtain the eigenvectors corresponding to the top eigenvalues. Every eigenvector $\mathbf{u}_i$ is an $(m+n)$-dimensional vector where the entries (sample indices) with significant values greater than a threshold $\rho$ are clustered as mode $i$. In our visualization, we show top-$r$ images with the maximum entry value on the shown top eigenvectors as the top novel modes.     

\noindent \textbf{Novel modes between real datasets.} Based on the proposed method, we visualized the novel modes samples' across content-similar datasets (AFHQ, ImageNet-dogs) and ({AFHQ, Wildlife}). Figure~\ref{fig:visual_realmode} visualizes the identified samples from the top three novel modes (principal eigenvectors). We observed that the detected modes exhibit semantically meaningful picture types of novel wildlife types missing in the ImageNet-dogs samples and novel docile "cat" and "dog" types compared to the Wildlife dataset. We did not find such samples when searching for these image types in the reference datasets. We postpone the presentation of similar results on other dataset pairs and CLIP and DINOv2 embeddings to the Appendix. \\
\textbf{Novel modes between standard generative models.} We analyzed FFHQ-trained models: GAN-based InsGen, StyleGAN2, StyleGAN-XL, diffusion-based LDM, and VAE-based VDVAE. Figure \ref{fig:visual_genmode} illustrates samples from the detected top novel mode between different pairs of generative models. For example, we observed that InsGen has more novelty in "people wearing sunglasses" than LDM, and StyleGAN2 has more novelty in "kids" than VDVAE. We present and discuss more visualizations for other pairs of generative models in Appendix A.2. According to Table \ref{tab:benchmark_novelty}, considering the averaged KEN scores over all ${5\choose 2}$ pairs, InsGen obtained the maximum averaged-KEN among the tested generative models in the FFHQ case. \\
\textbf{Detection of missing sample types of generative models.} To detect missing modes of the generators, we used the larger value $\eta=10$ for parameter $\eta$ in the computation of $K_{\mathbf{X}|\eta \mathbf{Y}}$. For example, our experimental results suggest the sample types "Microphone", "Round hat", and "Black uniform hat" to be not well-expressed in LDM. The visualization of our numerical results is postponed to the Appendix. In the Appendix, we will also present the applications of our spectral method for conditionally generating novel-mode samples and benchmarking model fitness.\vspace{-2mm}



\section{Conclusion}
In this paper, we proposed a spectral method for the evaluation of the novel modes in a mixture distribution $P$ which are expressed more frequently than in a reference distribution $Q$. We defined the KEN score to measure the entropy of the novel modes and tested the evaluation method on benchmark synthetic and image datasets. We note that our numerical evaluation focused on computer vision settings, and its extension to language models will be an interesting future direction. Also, characterizing tight statistical and computational complexity bounds for the novelty evaluation method will be a related topic for future exploration.  \vspace{-2mm}  

\section{Limitations}
Similar to other evaluation methods for image-based generative models, the results of KEN novelty evaluation are influenced by the choice of embedding, which may lead to biased results under ImageNet pre-trained models, such as the standard Inception-V3. Our work mainly focused on introducing and developing the kernel method for KEN novelty evaluation, and we leave a detailed analysis of the role of embedding in the novelty assessment, similar to the analysis in \cite{kyn2023,stein2023exposing} for quality and diversity metrics, for future studies. Furthermore, the spectral algorithm for KEN evaluation requires eigendecomposition of an $(n+m)\times (n+m)$ kernel matrix for $n,\, m $ test and reference samples, whose computational complexity $O\bigl((n+m)^3\bigr)$  will remain a barrier towards applying the framework to large sample sizes needed for large-scale datasets e.g. ImageNet. Exploring scalable extensions of the KEN framework is an interesting future direction.


\section*{Acknowledgements}
The work of Farzan Farnia is partially supported by a grant from the
Research Grants Council of the Hong Kong Special Administrative Region, China, Project 14209920, and is partially supported by a CUHK Direct Research Grant with CUHK Project No. 4055164. 
The work of Cheuk Ting Li is partially supported by two grants from the Research Grants Council of the Hong Kong Special Administrative Region, China [Project No.s: CUHK 24205621 (ECS), CUHK 14209823 (GRF)].

{
\bibliography{example_paper}
\bibliographystyle{unsrt}
}

\newpage
\appendix
\onecolumn
\section{Appendix}
\subsection{Proofs}
\subsubsection{Proof of Theorem 1}
To prove the theorem, note that every mode variable $\mathbf{X}_i\sim P_i$ can be written as $\mathbf{X}_i = \boldsymbol{\mu}_i + \mathbf{V}_i$ where $\mathbf{V}_i$ is a zero-mean random vector satisfying a bounded second-order moment $\mathbb{E}\bigl[\Vert\mathbf{V}_i\Vert^2_2\bigr]\le\sigma_i^2$. 
Then, we can decompose the kernel covariance  matrix $C_{\mathbf{X}}$ into the following two terms:
\begin{align*}
    C_{\mathbf{X}} \, 
    =\, \sum_{i=1}^k \Bigl[\omega_i \phi(\boldsymbol{\mu}_i)\phi(\boldsymbol{\mu}_i)^\top\Bigr] + \sum_{i=1}^k \Bigl[\omega_i \Bigr( \mathbb{E}\bigl[\phi(\boldsymbol{\mu}_i+\mathbf{V}_i)\phi(\boldsymbol{\mu}_i+\mathbf{V}_i)^\top \bigr]-\phi(\boldsymbol{\mu}_i)\phi(\boldsymbol{\mu}_i)^\top \Bigr) \Bigr].
\end{align*}
Therefore, we can write
\begin{align*}
    \Bigl\Vert C_{\mathbf{X}} - \sum_{i=1}^k \omega_i \phi(\boldsymbol{\mu}_i)\phi(\boldsymbol{\mu}_i)^\top \Bigr\Vert^2_F  
    =\, &\Bigl\Vert \sum_{i=1}^k \biggl[\omega_i \Bigr( \mathbb{E}\bigl[\phi(\boldsymbol{\mu}_i+\mathbf{V}_i)\phi(\boldsymbol{\mu}_i+\mathbf{V}_i)^\top \bigr]-\phi(\boldsymbol{\mu}_i)\phi(\boldsymbol{\mu}_i)^\top \Bigr) \biggr] \Bigr\Vert^2_F \\
    \stackrel{(a)}{\le} \, & \sum_{i=1}^k \biggl[\omega_i \Bigl\Vert \mathbb{E}\bigl[\phi(\boldsymbol{\mu}_i+\mathbf{V}_i)\phi(\boldsymbol{\mu}_i+\mathbf{V}_i)^\top \bigr]-\phi(\boldsymbol{\mu}_i)\phi(\boldsymbol{\mu}_i)^\top \Bigr\Vert^2_F \biggr] \\
    = \, & \sum_{i=1}^k \biggl[\omega_i \Bigl\Vert \mathbb{E}\bigl[\phi(\boldsymbol{\mu}_i+\mathbf{V}_i)\phi(\boldsymbol{\mu}_i+\mathbf{V}_i)^\top -\phi(\boldsymbol{\mu}_i)\phi(\boldsymbol{\mu}_i)^\top \bigr] \Bigr\Vert^2_F \biggr] \\
    \stackrel{(b)}{\le} \, & \sum_{i=1}^k \omega_i  \mathbb{E}\Bigl[\Bigl\Vert\phi(\boldsymbol{\mu}_i+\mathbf{V}_i)\phi(\boldsymbol{\mu}_i+\mathbf{V}_i)^\top -\phi(\boldsymbol{\mu}_i)\phi(\boldsymbol{\mu}_i)^\top\Bigr\Vert^2_F \Bigr] 
    \\
    \stackrel{(c)}{=} \, & \sum_{i=1}^k \omega_i  \mathbb{E}\Bigl[2-2\bigl(\phi(\boldsymbol{\mu}_i)^\top\phi(\boldsymbol{\mu}_i+\mathbf{V}_i)\bigr)^2 \Bigr] 
    \\
    = \, & \sum_{i=1}^k 2\omega_i  \mathbb{E}\Bigl[1-\exp\bigl(-\frac{\Vert \mathbf{V}_i\Vert_2^2}{\sigma^2}\bigr) \Bigr] \\
   \stackrel{(d)}{\le} \, & \sum_{i=1}^k 2\omega_i  \Bigl(1-\exp\bigl(-\frac{\mathbb{E}\bigl[\Vert \mathbf{V}_i\Vert_2^2\bigr]}{\sigma^2}\bigr) \Bigr) \\
   \le \, & \sum_{i=1}^k 2\omega_i  \Bigl(1-\exp\bigl(-\frac{\sigma_i^2}{\sigma^2}\bigr) \Bigr) \\
   \stackrel{(e)}{\le} \, & 2\sum_{i=1}^k  \omega_i\frac{\sigma_i^2}{\sigma^2} 
\end{align*}
In the above, (a) and (b) follow from Jensen's inequality applied to the convex Frobenius-norm-squared function.
(c) holds because given the unit-norm vectors $\mathbf{a}=\phi(\boldsymbol{\mu}_i+\mathbf{V}_i)$ and $\mathbf{b}=\phi(\boldsymbol{\mu}_i)$ the following holds $$\bigl\Vert \phi(\boldsymbol{\mu}_i+\mathbf{V}_i)\phi(\boldsymbol{\mu}_i+\mathbf{V}_i)^\top -\phi(\boldsymbol{\mu}_i)\phi(\boldsymbol{\mu}_i)^\top \bigr\Vert^2_F = \Vert \mathbf{a}\mathbf{a}^\top - \mathbf{b}\mathbf{b}^\top\Vert^2_F = \Vert \mathbf{a}\Vert^4 + \Vert \mathbf{b}\Vert^4 - 2(\mathbf{a}^\top\mathbf{b})^2 = 2- 2(\mathbf{a}^\top\mathbf{b})^2.$$
Finally, (d) follows from Jensen's inequality for the concave function $t(z)=1-\exp(-z)$, and (e) holds because of the inequality $1-e^{-t}\le t$ for every $t\in\mathbb{R}$. Next, we create the following orthogonal basis consisting of vectors $\mathbf{u}_1,\ldots , \mathbf{u}_k$ of the span of the $k$ unit-norm vectors $\phi(\boldsymbol{\mu}_1),\ldots , \phi(\boldsymbol{\mu}_k)$ as follows: We choose $\mathbf{u}_1=\phi(\boldsymbol{\mu}_1)$, and for every $2\le i\le k$ we construct $\mathbf{u}_i$ as
\begin{align*}
    \mathbf{u}_i\, :=\, &\phi(\boldsymbol{\mu}_i) - \sum_{j=1}^{i-1} \langle \phi(\boldsymbol{\mu}_i) ,\mathbf{u}_j\rangle \mathbf{u}_j. 
\end{align*}
Therefore, we will have:
\begin{align*}
    \Bigl\Vert \sum_{i=1}^k \omega_i \phi(\boldsymbol{\mu}_i)\phi(\boldsymbol{\mu}_i)^\top - \sum_{i=1}^k \omega_i \mathbf{u}_i\mathbf{u}_i^\top \Bigr\Vert^2_F \, &= \, \Bigl\Vert \sum_{i=1}^k \omega_i\Bigl( \phi(\boldsymbol{\mu}_i)\phi(\boldsymbol{\mu}_i)^\top - \mathbf{u}_i\mathbf{u}_i^\top \Bigr)\Bigr\Vert^2_F \\
    &\stackrel{(f)}{\le} \,  \sum_{i=1}^k \omega_i\Bigl\Vert \phi(\boldsymbol{\mu}_i)\phi(\boldsymbol{\mu}_i)^\top - \mathbf{u}_i\mathbf{u}_i^\top \Bigr\Vert^2_F \\
    &\stackrel{(g)}{=} \,  \sum_{i=1}^k \omega_i\Bigl( 1+ \Vert \mathbf{u}_i\Vert^4-2\bigl(\mathbf{u}_i^\top\phi(\boldsymbol{\mu}_i)\bigr)^2 \Bigr) \\
    &\le \,  \sum_{i=1}^k \omega_i\Bigl( 2-2\bigl(\mathbf{u}_i^\top\phi(\boldsymbol{\mu}_i)\bigr)^2 \Bigr)
    \\
    &= \,  \sum_{i=1}^k 2\omega_i\Bigl( 1+\mathbf{u}_i^\top\phi(\boldsymbol{\mu}_i) \Bigr)\Bigl( 1-\mathbf{u}_i^\top\phi(\boldsymbol{\mu}_i) \Bigr)\\
    &\stackrel{(h)}{\le} \,  \sum_{i=1}^k\sum_{j=1}^{i-1} 4\omega_i\exp\Bigl(\frac{-\Vert\boldsymbol{\mu}_i - \boldsymbol{\mu}_j \Vert^2_2}{\sigma^2} \Bigr). 
\end{align*}
In the above, (f) comes from the application of Jensen's inequality for the convex Frobenius norm-squared function. (g) follows because of the same reason as for item (c). (h) holds because $1+ \mathbf{u}_i^\top\phi(\boldsymbol{\mu}_i) \le 2$ and
\begin{align*}
    \mathbf{u}_i^\top\phi(\boldsymbol{\mu}_i) = 1- \sum_{j=1}^{i-1} \langle \phi(\boldsymbol{\mu}_i) ,\mathbf{u}_j\rangle^2 \ge 1- \sum_{j=1}^{i-1} \exp\Bigl(\frac{-\Vert \boldsymbol{\mu}_i- \boldsymbol{\mu}_j\Vert^2_2}{\sigma^2}\Bigr).
\end{align*}
Since for every two matrices $A,\,B$ we have $\Vert A + B\Vert_F^2 \le 2\Vert A\Vert_F^2 + 2\Vert B\Vert_F^2$, we can combine the previous shown inequalities to obtain
\begin{align*}
     \Bigl\Vert C_{\mathbf{X}} - \sum_{i=1}^k \omega_i \mathbf{u}_i\mathbf{u}_i^\top \Bigr\Vert^2_F \: &\le \:  2\Bigl\Vert C_{\mathbf{X}} - \sum_{i=1}^k \omega_i \phi(\boldsymbol{\mu}_i)\phi(\boldsymbol{\mu}_i)^\top \Bigr\Vert^2_F + 2\Bigl\Vert \sum_{i=1}^k \omega_i \phi(\boldsymbol{\mu}_i)\phi(\boldsymbol{\mu}_i)^\top - \sum_{i=1}^k \omega_i \mathbf{u}_i\mathbf{u}_i^\top \Bigr\Vert^2_F \\
     &\le \: 4\sum_{i=1}^k  \frac{\omega_i\sigma_i^2}{\sigma^2}   +  8\sum_{i=1}^k\sum_{j=1}^{i-1} \omega_i\exp\Bigl(\frac{-\Vert\boldsymbol{\mu}_i - \boldsymbol{\mu}_j \Vert^2_2}{\sigma^2} \Bigr)
\end{align*}
Since we know that $\Vert \mathbf{u}_i\Vert^2\omega_i$ for $i=1,\ldots,k$ are the eigenvalues of $\sum_{i=1}^k \omega_i \mathbf{u}_i\mathbf{u}_i^\top$ where $1- 2\sum_{j=1}^{i-1}\exp\Bigl(\frac{-\Vert\boldsymbol{\mu}_i - \boldsymbol{\mu}_j \Vert^2_2}{\sigma^2} \Bigr)\le\Vert \mathbf{u}_i\Vert^2 \le 1$, then the eigenspectrum stability bound in \cite{hoffman2003variation} implies that for the top $k$ eigenvalues of $C_{\mathbf{X}}$, denoted by $\lambda_1,\ldots ,\lambda_k$, we will have
\begin{align*}
    \sum_{i=1}^k \bigl(\lambda_i -\Vert \mathbf{u}_i\Vert^2\omega_i\bigr)^2 \le 4\sum_{i=1}^k  \frac{\omega_i\sigma_i^2}{\sigma^2} +  8\sum_{i=1}^k\sum_{j=1}^{i-1} \omega_i\exp\Bigl(\frac{-\Vert\boldsymbol{\mu}_i - \boldsymbol{\mu}_j \Vert^2_2}{\sigma^2} \Bigr)
\end{align*}
Therefore, since $\bigl(\lambda_i -\Vert \mathbf{u}_i\Vert^2\omega_i\bigr)^2 \le \bigl(\lambda_i -\omega_i\bigr)^2 + 2(1-\Vert \mathbf{u}_i\Vert^2)\omega_i$, we obtain the following which completes the proof:
\begin{align*}
    \sum_{i=1}^k \bigl(\lambda_i -\omega_i\bigr)^2 \le 4\sum_{i=1}^k  \frac{\omega_i\sigma_i^2}{\sigma^2}  +  16\sum_{i=1}^k\sum_{j=1}^{i-1} \omega_i\exp\Bigl(\frac{-\Vert\boldsymbol{\mu}_i - \boldsymbol{\mu}_j \Vert^2_2}{\sigma^2} \Bigr).
\end{align*}
\subsubsection{Proof of Theorem 2}
To show the theorem, we first follow Theorem 1's proof where we showed that:
\begin{align*}
    \Bigl\Vert C_{\mathbf{Y}} - \sum_{i=1}^k \gamma_i \phi(\boldsymbol{\mu}'_i)\phi(\boldsymbol{\mu}'_i)^\top \Bigr\Vert^2_F  \,
   \le \, 2\sum_{i=1}^k  \frac{\gamma_i\sigma_i^2}{\sigma^2}
\end{align*}
Next, we attempt to bound the norm difference between $C_{\mathbf{X}}$ and $ \sum_{i=1}^k \omega_i \phi(\boldsymbol{\mu}'_i)\phi(\boldsymbol{\mu}'_i)^\top$:

\begin{align*}
    \Bigl\Vert C_{\mathbf{X}} - \sum_{i=1}^k \omega_i \phi(\boldsymbol{\mu}'_i)\phi(\boldsymbol{\mu}'_i)^\top \Bigr\Vert^2_F  
    =\, &\Bigl\Vert \sum_{i=1}^k \biggl[\omega_i \Bigr( \mathbb{E}\bigl[\phi(\boldsymbol{\mu}_i+\mathbf{V}_i)\phi(\boldsymbol{\mu}_i+\mathbf{V}_i)^\top \bigr]-\phi(\boldsymbol{\mu}'_i)\phi(\boldsymbol{\mu}'_i)^\top \Bigr) \biggr] \Bigr\Vert^2_F \\
    \stackrel{(a)}{\le} \, & \sum_{i=1}^k \biggl[\omega_i \Bigl\Vert \mathbb{E}\bigl[\phi(\boldsymbol{\mu}_i+\mathbf{V}_i)\phi(\boldsymbol{\mu}_i+\mathbf{V}_i)^\top \bigr]-\phi(\boldsymbol{\mu}'_i)\phi(\boldsymbol{\mu}'_i)^\top \Bigr\Vert^2_F \biggr] \\
    = \, & \sum_{i=1}^k \biggl[\omega_i \Bigl\Vert \mathbb{E}\bigl[\phi(\boldsymbol{\mu}_i+\mathbf{V}_i)\phi(\boldsymbol{\mu}_i+\mathbf{V}_i)^\top -\phi(\boldsymbol{\mu}'_i)\phi(\boldsymbol{\mu}'_i)^\top \bigr] \Bigr\Vert^2_F \biggr] \\
    \stackrel{(b)}{\le} \, & \sum_{i=1}^k \omega_i  \mathbb{E}\Bigl[\Bigl\Vert\phi(\boldsymbol{\mu}_i+\mathbf{V}_i)\phi(\boldsymbol{\mu}_i+\mathbf{V}_i)^\top -\phi(\boldsymbol{\mu}'_i)\phi(\boldsymbol{\mu}'_i)^\top\Bigr\Vert^2_F \Bigr] 
    \\
    \stackrel{(c)}{=} \, & \sum_{i=1}^k \omega_i  \mathbb{E}\Bigl[2-2\bigl(\phi(\boldsymbol{\mu}'_i)^\top\phi(\boldsymbol{\mu}_i+\mathbf{V}_i)\bigr)^2 \Bigr] 
    \\
    = \, & \sum_{i=1}^k 2\omega_i  \mathbb{E}\Bigl[1-\exp\bigl(\frac{-\Vert \mathbf{V}_i + \boldsymbol{\delta}_i\Vert_2^2}{\sigma^2}\bigr) \Bigr] \\
    \stackrel{(d)}{\le} \, & \sum_{i=1}^k 2\omega_i  \Bigl(1-\exp\bigl(\frac{-\mathbb{E}\bigl[\Vert \mathbf{V}_i + \boldsymbol{\delta}_i\Vert_2^2\bigr]}{\sigma^2}\bigr) \Bigr) \\
    \stackrel{(e)}{=} \, & \sum_{i=1}^k 2\omega_i  \Bigl(1-\exp\bigl(-\frac{\mathbb{E}\bigl[\Vert \mathbf{V}_i\Vert^2_2\bigr] + \Vert\boldsymbol{\delta}_i\Vert_2^2}{\sigma^2}\bigr) \Bigr) \\   
     \le \, & \sum_{i=1}^k 2\omega_i  \Bigl(1-\exp\bigl(-\frac{\sigma_i^2 + \Vert\boldsymbol{\delta}_i\Vert_2^2}{\sigma^2}\bigr) \Bigr) \\
     \stackrel{(f)}{\le} \, & 2\sum_{i=1}^k   \frac{\omega_i\bigl(\sigma_i^2 + \Vert\boldsymbol{\delta}_i\Vert_2^2\bigr)}{\sigma^2}
\end{align*}
Note that in the above (a), (b), (c), (d), and (f) hold for the same reason as the same-numbered items hold in the proof of Theorem 1. Also, (e) holds because $\mathbb{E}[\mathbf{V}_i]=\mathbf{0}$. 

Then, since for every matrices $A,B$ we have $\Vert A+B\Vert_F^2\le 2\Vert A \Vert^2_F + 2\Vert B \Vert^2_F$, we can combine the above two parts to show:
\begin{align*}
    &\Bigl\Vert \Bigl(C_{\mathbf{X}} - \eta C_{\mathbf{Y}}\Bigr) - \sum_{i=1}^k (\omega_i-\eta\gamma_i)  \phi(\boldsymbol{\mu}'_i)\phi(\boldsymbol{\mu}'_i)^\top  \Bigr\Vert^2_F \\
    =\, &\Bigl\Vert \Bigl(C_{\mathbf{X}} - \sum_{i=1}^k \omega_i \phi(\boldsymbol{\mu}'_i)\phi(\boldsymbol{\mu}'_i)^\top\Bigr) - \eta \Bigl( C_{\mathbf{Y}} -  \sum_{i=1}^k \gamma_i \phi(\boldsymbol{\mu}'_i)\phi(\boldsymbol{\mu}'_i)^\top \Bigr)\Bigr\Vert^2_F \\
   \le \, &2\Bigl\Vert C_{\mathbf{X}} - \sum_{i=1}^k \omega_i \phi(\boldsymbol{\mu}'_i)\phi(\boldsymbol{\mu}'_i)^\top\Bigr\Vert_F^2 +2 \eta^2 \Bigl\Vert C_{\mathbf{Y}} -  \sum_{i=1}^k \gamma_i \phi(\boldsymbol{\mu}'_i)\phi(\boldsymbol{\mu}'_i)^\top \Bigr)\Bigr\Vert^2_F \\
    \,
   \le \, & 4\sum_{i=1}^k   \frac{\omega_i\bigl(\sigma_i^2 + \Vert\boldsymbol{\delta}_i\Vert_2^2\bigr)+\eta^2\gamma_i \sigma_i^2}{\sigma^2}\\
    \,
   = \, & 4\sum_{i=1}^k   \frac{\bigl(\omega_i+\eta^2\gamma_i\bigr)\sigma_i^2 +\omega_i \Vert\boldsymbol{\delta}_i\Vert_2^2}{\sigma^2}
\end{align*}
Next, we create an orthogonal basis consisting of vectors $\mathbf{u}_1,\ldots , \mathbf{u}_t$ of the span of the $t$ unit-norm vectors $\phi(\boldsymbol{\mu}'_1),\ldots , \phi(\boldsymbol{\mu}'_t)$ as follows where for every $1\le i\le t$ we construct $\mathbf{u}_i$ as
\begin{align*}
    \mathbf{u}_i\, :=\, &\phi(\boldsymbol{\mu}_i) - \sum_{j=1}^{i-1} \langle \phi(\boldsymbol{\mu}_i) ,\mathbf{u}_j\rangle \mathbf{u}_j 
\end{align*}
As a result, we can show:
\begin{align*}
    \Bigl\Vert \sum_{i=1}^k (\omega_i-\eta\gamma_i) \phi(\boldsymbol{\mu}'_i)\phi(\boldsymbol{\mu}'_i)^\top - \sum_{i=1}^k (\omega_i-\eta\gamma_i) \mathbf{u}_i\mathbf{u}_i^\top \Bigr\Vert^2_F \, &= \, \Bigl\Vert \sum_{i=1}^k (\omega_i-\eta\gamma_i)\Bigl( \phi(\boldsymbol{\mu}'_i)\phi(\boldsymbol{\mu}'_i)^\top - \mathbf{u}_i\mathbf{u}_i^\top \Bigr)\Bigr\Vert^2_F \\
    &\stackrel{}{\le} \,  \sum_{i=1}^k (1+\eta)(\omega_i+\eta\gamma_i)\Bigl\Vert \phi(\boldsymbol{\mu}'_i)\phi(\boldsymbol{\mu}'_i)^\top - \mathbf{u}_i\mathbf{u}_i^\top \Bigr\Vert^2_F \\
    &\stackrel{}{=} \,  \sum_{i=1}^k (1+\eta)(\omega_i+\eta\gamma_i)\Bigl( 1+ \Vert \mathbf{u}_i\Vert^4-2\bigl(\mathbf{u}_i^\top\phi(\boldsymbol{\mu}'_i)\bigr)^2 \Bigr) \\
    &\stackrel{}{\le} \,  \sum_{i=1}^k\sum_{j=1}^{i-1} 4(1+\eta)(\omega_i+\eta\gamma_i)\exp\Bigl(\frac{-\Vert\boldsymbol{\mu}'_i - \boldsymbol{\mu}'_j \Vert^2_2}{\sigma^2} \Bigr). 
\end{align*}
Therefore, we can combine the above results to show:
\begin{align*}
    &\Bigl\Vert \bigl(C_{\mathbf{X}}-\eta C_{\mathbf{Y}}\bigr) - \sum_{i=1}^k (\omega_i-\eta\gamma_i) \mathbf{u}_i\mathbf{u}_i^\top \Bigr\Vert^2_F \\
    \stackrel{}{\le} \, &8\sum_{i=1}^k   \frac{\bigl(\omega_i+\eta^2\gamma_i\bigr)\sigma_i^2 +\omega_i \Vert\boldsymbol{\delta}_i\Vert_2^2}{\sigma^2} \\
    &\quad + 8(1+\eta)\sum_{i=1}^k\sum_{j=1}^{i-1} (\omega_i+\eta\gamma_i)\exp\Bigl(\frac{-\Vert\boldsymbol{\mu}'_i - \boldsymbol{\mu}'_j \Vert^2_2}{\sigma^2} \Bigr). 
\end{align*}
Since we know that $\Vert \mathbf{u}_i\Vert^2(\omega_i-\eta \gamma_i)$ for $i=1,\ldots,t$ are the eigenvalues of $\sum_{i=1}^k (\omega_i-\eta \gamma_i) \mathbf{u}_i\mathbf{u}_i^\top$ where $1- 2\sum_{j=1}^{i-1}\exp\Bigl(\frac{-\Vert\boldsymbol{\mu}'_i - \boldsymbol{\mu}_j \Vert^2_2}{\sigma^2} \Bigr)\le\Vert \mathbf{u}_i\Vert^2 \le 1$, then the eigenspectrum stability bound in \cite{hoffman2003variation} shows that for the top $k$ eigenvalues of $C_{\mathbf{X}} - \eta C_{\mathbf{Y}}$, denoted by $\lambda_1\ge \ldots \ge \lambda_{k}$, we will have
\begin{align*}
    &\sum_{i=1}^{k} \bigl(\lambda_i -\Vert \mathbf{u}_i\Vert^2(\omega_{i}-\eta \gamma_{i} )\bigr)^2 \\
    \stackrel{}{\le} \, &8\sum_{i=1}^k   \frac{\bigl(\omega_i+\eta^2\gamma_i\bigr)\sigma_i^2 +\omega_i \Vert\boldsymbol{\delta}_i\Vert_2^2}{\sigma^2} \, +\, 8(1+\eta)\sum_{i=1}^k\sum_{j=1}^{i-1} (\omega_i+\eta\gamma_i)\exp\Bigl(\frac{-\Vert\boldsymbol{\mu}'_i - \boldsymbol{\mu}'_j \Vert^2_2}{\sigma^2} \Bigr). 
\end{align*}
As a consequence, since $\bigl(\lambda_i -\Vert \mathbf{u}_i\Vert^2(\omega_{i}-\eta \gamma_{i} )\bigr)^2 \le \bigl(\lambda_i -(\omega_{i}-\eta \gamma_{i} )\bigr)^2 + 2(1-\Vert \mathbf{u}_i\Vert^2)\max\{\omega_{i}-\eta \gamma_{i}  , 0\}$ and $\mathrm{ReLU}(z)=\max\{z,0\}$ is a 1-Lipschitz function, we obtain the following which finishes the proof:
\begin{align*}
    &\sum_{i=1}^k \bigl(\max\{\lambda_i,0\} -\max\{\omega_{i}-\eta \gamma_{i} ,0\}\bigr)^2 \\
    \stackrel{}{\le} \, &8\sum_{i=1}^k   \frac{\bigl(\omega_i+\eta^2\gamma_i\bigr)\sigma_i^2 +\omega_i \Vert\boldsymbol{\delta}_i\Vert_2^2}{\sigma^2} \, +\, 16(1+\eta)\sum_{i=1}^k\sum_{j=1}^{i-1} (\omega_i+\eta\gamma_i)\exp\Bigl(\frac{-\Vert\boldsymbol{\mu}'_i - \boldsymbol{\mu}'_j \Vert^2_2}{\sigma^2} \Bigr)  
\end{align*}
\subsubsection{Proof of Theorem 3}
We note that given the empirical kernel feature matrices $\Phi_{\mathbf{X}}\in\mathbb{R}^{n\times s}$ and $\Phi_{\mathbf{X}}\in\mathbb{R}^{m\times s}$, we can write 
\begin{equation*}
    \widehat{C}_{\mathbf{X}} = \frac{1}{n}\Phi_{\mathbf{X}}^\top\Phi_{\mathbf{X}}, \quad \widehat{C}_{\mathbf{Y}} = \frac{1}{m}\Phi_{\mathbf{Y}}^\top\Phi_{\mathbf{Y}}. 
\end{equation*}
Therefore, defining $\tilde{\Phi}_{\mathbf{X}}= \frac{1}{\sqrt{n}}\Phi_{\mathbf{X}}$ and $\tilde{\Phi}_{\mathbf{Y}}= \frac{1}{\sqrt{m}}\Phi_{\mathbf{Y}}$, we can rewrite the definition of the $\eta$-differential kernel covariance matrix as
\begin{align*}
   \widehat{C}_{\mathbf{X}}  - \eta \widehat{C}_{\mathbf{Y}} &=  \tilde{\Phi}_{\mathbf{X}}^\top \tilde{\Phi}_{\mathbf{X}} - \eta \tilde{\Phi}_{\mathbf{Y}}^\top \tilde{\Phi}_{\mathbf{Y}} \\
   &= \begin{bmatrix} \tilde{\Phi}_{\mathbf{X}} \vspace{1mm} \\ \sqrt{\eta}\tilde{\Phi}_{\mathbf{Y}} \end{bmatrix}^\top \begin{bmatrix} \tilde{\Phi}_{\mathbf{X}} \vspace{1mm} \\
     -\sqrt{\eta}\tilde{\Phi}_{\mathbf{Y}} \end{bmatrix}.
\end{align*}
Defining $A = \begin{bmatrix} \tilde{\Phi}_{\mathbf{X}} \hspace{1mm} \\ \sqrt{\eta}\tilde{\Phi}_{\mathbf{Y}} \end{bmatrix}$ and $B=\begin{bmatrix} \tilde{\Phi}_{\mathbf{X}} \vspace{1mm} \\
     -\sqrt{\eta}\tilde{\Phi}_{\mathbf{Y}} \end{bmatrix}$, we use the property that $A^\top B$ and $BA^\top$ share the same non-zero eigenvalues, because if for $\lambda\neq 0$ and $\mathbf{v}$ we have $A^\top B \mathbf{v}= \lambda \mathbf{v}$, then for $\mathbf{u}= B\mathbf{v}$ we have $BA^\top  \mathbf{u}= \lambda \mathbf{u}$. Therefore, the non-zero eigenvalues of the $\eta$-differential kernel covariance matrix $\Lambda_{\mathbf{X}|\eta \mathbf{Y}} = \widehat{C}_{\mathbf{X}}  - \eta \widehat{C}_{\mathbf{Y}}$ will be the same as the non-zero eigenevalues of 
     \begin{align*}
         \begin{bmatrix} \tilde{\Phi}_{\mathbf{X}} \vspace{1mm} \\
     -\sqrt{\eta}\tilde{\Phi}_{\mathbf{Y}} \end{bmatrix} \begin{bmatrix} \tilde{\Phi}_{\mathbf{X}} \vspace{1mm} \\ \sqrt{\eta}\tilde{\Phi}_{\mathbf{Y}} \end{bmatrix}^\top \, &=\, \begin{bmatrix} \tilde{\Phi}_{\mathbf{X}}\tilde{\Phi}_{\mathbf{X}}^\top\hspace{1mm} & \sqrt{\eta}\tilde{\Phi}_{\mathbf{X}}\tilde{\Phi}_{\mathbf{Y}}^\top   \vspace{1mm} \\
    -\sqrt{\eta}\tilde{\Phi}_{\mathbf{Y}}\tilde{\Phi}_{\mathbf{X}}^\top \hspace{1mm} & -\eta \tilde{\Phi}_{\mathbf{Y}}\tilde{\Phi}_{\mathbf{Y}}^\top \end{bmatrix} \\
    &= \, \begin{bmatrix} K_{\mathbf{X}\mathbf{X}}\hspace{1mm} & \sqrt{\eta}\, K_{\mathbf{X}\mathbf{Y}}\hspace{1mm}\vspace{1mm} \\ -\sqrt{\eta}\, K^\top_{\mathbf{X}\mathbf{Y}}\hspace{1mm} & -\eta\, K_{\mathbf{Y}\mathbf{Y}}\hspace{1mm}\vspace{1mm}  \end{bmatrix}\\ \, &=\, K_{\mathbf{X}|\eta \mathbf{Y}}.
     \end{align*}
In addition, given every eigenvector $\mathbf{v}$ of $K_{\mathbf{X}|\eta \mathbf{Y}}$, the vector $\mathbf{u} = A^\top \mathbf{v} = \begin{bmatrix} \tilde{\Phi}_{\mathbf{X}} \hspace{1mm} \\ \sqrt{\eta}\tilde{\Phi}_{\mathbf{Y}} \end{bmatrix}^\top \mathbf{v}$ will be an eigenvector of $\Lambda_{\mathbf{X}|\eta \mathbf{Y}} = \widehat{C}_{\mathbf{X}}  - \eta \widehat{C}_{\mathbf{Y}}$, which will be
\begin{equation*}
    \begin{bmatrix} \tilde{\Phi}_{\mathbf{X}} \hspace{1mm} \\ \sqrt{\eta}\tilde{\Phi}_{\mathbf{Y}} \end{bmatrix}^\top \mathbf{v} \, = \, \sum_{i=1}^n v_i \phi(\mathbf{x}_i) +  \sum_{j=1}^m \sqrt{\eta}v_{n+j} \phi(\mathbf{y}_j).
\end{equation*}
Therefore, the proof is complete.
\subsubsection{Proof of Theorem 4}
First, we note that $K_{\mathbf{X},\eta \mathbf{Y}}$ is a symmetric PSD matrix, because defining $A = \begin{bmatrix} \tilde{\Phi}_{\mathbf{X}} \hspace{1mm} \\ \sqrt{\eta}\tilde{\Phi}_{\mathbf{Y}} \end{bmatrix}$ where $\tilde{\Phi}_{\mathbf{X}}= \frac{1}{\sqrt{n}}\Phi_{\mathbf{X}}$ and $\tilde{\Phi}_{\mathbf{Y}}= \frac{1}{\sqrt{m}}\Phi_{\mathbf{Y}}$ we will have $K_{\mathbf{X},\eta \mathbf{Y}} = AA^\top$. Therefore, applying the Cholesky decomposition, we can find a $V\in\mathbb{R}^{(m+n)\times (m+n)}$ such that $K_{\mathbf{X},\eta \mathbf{Y}} = V^\top V$.

Next, we note the following identity given $D= \mathrm{diag}\{[\underbrace{+1,\ldots , +1}_{\text{\rm n times}},\underbrace{-1,\ldots , -1}_{\text{\rm m times}}]\}$:
\begin{align*}
    K_{\mathbf{X}|\eta \mathbf{Y}}\: &=\: \begin{bmatrix} K_{\mathbf{X}\mathbf{X}}\hspace{1.5mm} & \sqrt{\eta}\, K_{\mathbf{X}\mathbf{Y}}\hspace{1.5mm}\vspace{3mm} \\ -\sqrt{\eta}\, K^\top_{\mathbf{X}\mathbf{Y}}\hspace{1.5mm} & -\eta\, K_{\mathbf{Y}\mathbf{Y}}\hspace{1.5mm}\vspace{1mm}  \end{bmatrix} \\
    \, &=\: D\begin{bmatrix} K_{\mathbf{X}\mathbf{X}}\hspace{1.5mm} & \sqrt{\eta}\, K_{\mathbf{X}\mathbf{Y}}\hspace{1.5mm}\vspace{3mm} \\ \sqrt{\eta}\, K^\top_{\mathbf{X}\mathbf{Y}}\hspace{1.5mm} & \eta\, K_{\mathbf{Y}\mathbf{Y}}\hspace{1.5mm}\vspace{1mm}  \end{bmatrix} \\
    &=\: D K_{\mathbf{X},\eta \mathbf{Y}} \\
    &=\: D V^\top V. 
\end{align*}
However, we observe that based on the same argument in Theorem 2's proof, $D V^\top V$ and $ V D V^\top$ have the same non-zero eigenvalues. Therefore, the symmetric matrix $VDV^\top$ and $K_{\mathbf{X}|\eta \mathbf{Y}}$ share the same non-zero eigenvalues.

\subsection{Experimental Results}

\subsubsection{Applications of KEN Score}

\begin{figure*}
    \centering
    \includegraphics[width=\textwidth]{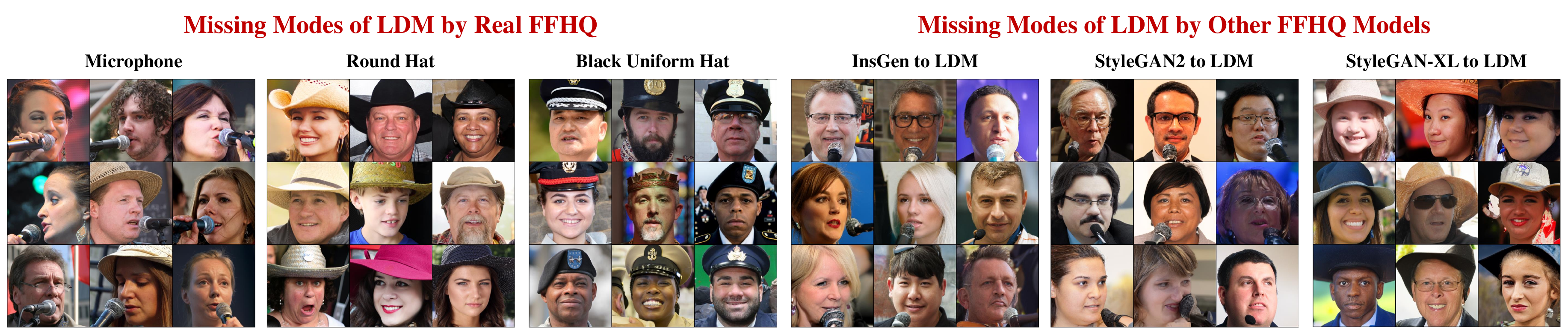}
    \vspace{-5mm}
    \caption{Missing modes of LDM in FFHQ. Training set and other generative models both capture similar missing modes of LDM with large $\eta=10$ in $K_{X|Y}$. Inception-V3 embedding is used.}
    \label{fig:visual_missingmode}
\end{figure*}

\begin{figure*}
    \centering
    \includegraphics[width=.7\textwidth]{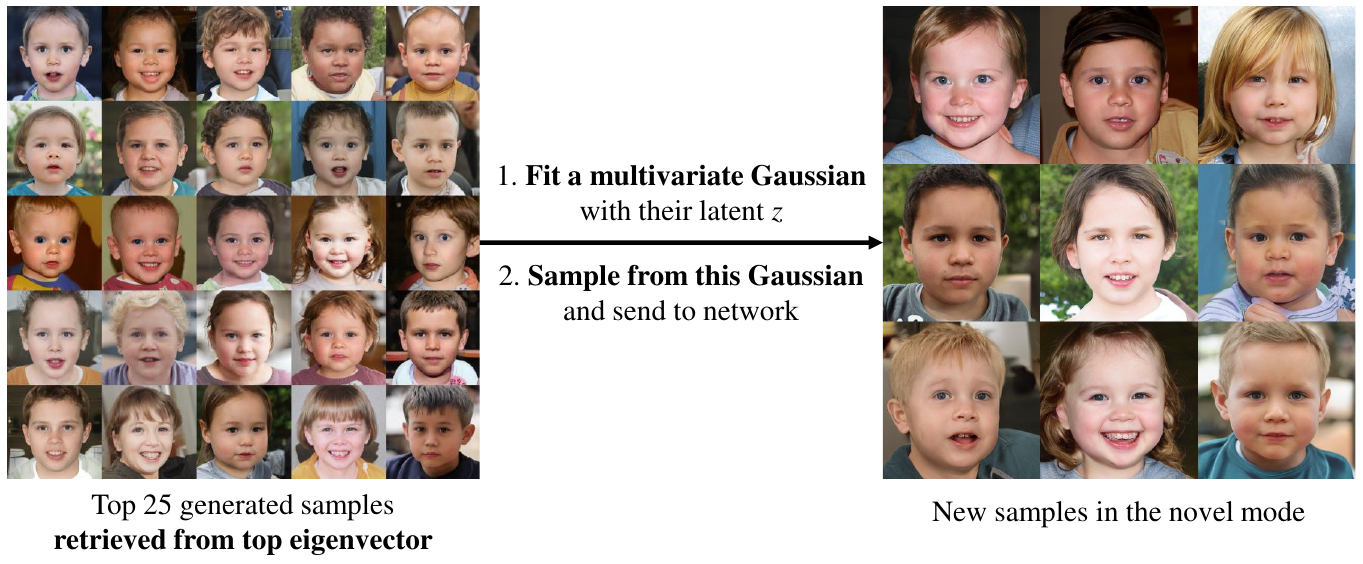}
    \caption{Generating new samples in a specific novel mode by fitting a Gaussian with samples from qualitative analysis. Inception-V3 embedding is used.}
    \label{fig:application_gen}
\end{figure*}

\textbf{Missing mode detection.} To enable missing mode detection, we can select a large enough $\eta$ for $K_{\mathbf{X}|\eta \mathbf{Y}}$. For example, according to Figure~\ref{fig:visual_missingmode}, the modes "Microphone", "Round hat", and "Black uniform hat" are found missing in LDM by its training set and other generative models.\\
\textbf{Specific novel mode generation.} The qualitative analysis can reveal most related samples of a novel mode. Therefore, we can retrieve the latent $z$ of these novel samples to fit a Gaussian. Then, we sample from this Gaussian to obtain new samples in the same novel mode. We put an example of specifically generating more FFHQ "kids" with StyleGAN-XL in Figure \ref{fig:application_gen}. \\
\textbf{Benchmarking mode novelty.} For a group of generative models with the same training set. We can evaluate the mode novelty between them. The average novelty of a generative model to others can be used for benchmarking. Table \ref{tab:benchmark_novelty} shows mode novelty between generative models trained on FFHQ. We observe that InsGen has the highest average novelty and VDVAE has the lowest average novelty in this group. \\
\textbf{Benchmarking fitness.} If we use generative models and their training sets as testing and reference distribution, our proposed KEN can be recognized as a divergence measurement. When two distributions are identical, their KEN evaluation will be 0. In Table \ref{tab:benchmark_fitness}, we observe KEN behave similarly with FID in ImageNet and CIFAR-10, except for GGAN, DCGAN, and WGAN in CIFAR-10.

\subsubsection{Extra Quantified Analysis of KEN Score}
\begin{figure*}
    \centering
    \includegraphics[width=\textwidth]{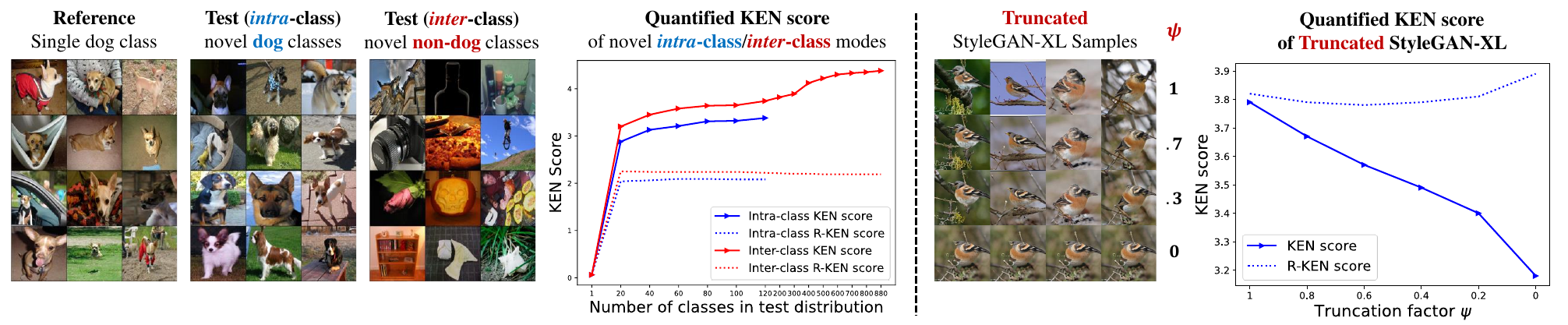}
    \vspace{-5mm}
    \caption{Quantified KEN score in real and generated distributions. \textbf{Left: }KEN score in ImageNet-1K. \emph{Intra}-class means similarity in taxonomy (e.g. Dogs with different breeds). \textbf{Right: }KEN score in truncated StyleGAN-XL. $\psi$ is truncation factor. $\psi=1$ reduces to normal StyleGAN-XL. \emph{"R-KEN"} means switching test and reference distributions. Inception-V3 embedding is used.}
    \label{fig:quantified}
\end{figure*}

\textbf{Distinct modes contain richer novelty.} To define similar modes, we extract 120 dog classes from ImageNet-1K. The remaining 880 classes are dog-excluded and represent distinct modes. We select a single dog class as the reference, other dog classes as novel \emph{intra}-class modes, and 880 dog-excluded classes as novel \emph{inter}-class modes. Figure \ref{fig:quantified} shows that adding novel modes to test distribution increases mode novelty. Meanwhile, the line chart in Figure \ref{fig:quantified} indicates \emph{inter}-class modes contain richer novelty than \emph{intra}-class modes since the red \emph{inter}-class line is higher. The reversed novelty lines remain flat, illustrating the asymmetric property. 

\textbf{Truncation trick decreases mode novelty.} Truncation trick \cite{marchesi2017megapixel, brock2018large} is a procedure sampling latent $z$ from a truncated normal to trade-off diversity for high-fidelity generated images. We observe this trick also reduces the KEN score of generative model in Figure \ref{fig:quantified}.

\subsubsection{Extra Experimental Results}
\begin{figure*}
    \centering
    \includegraphics[width=\textwidth]{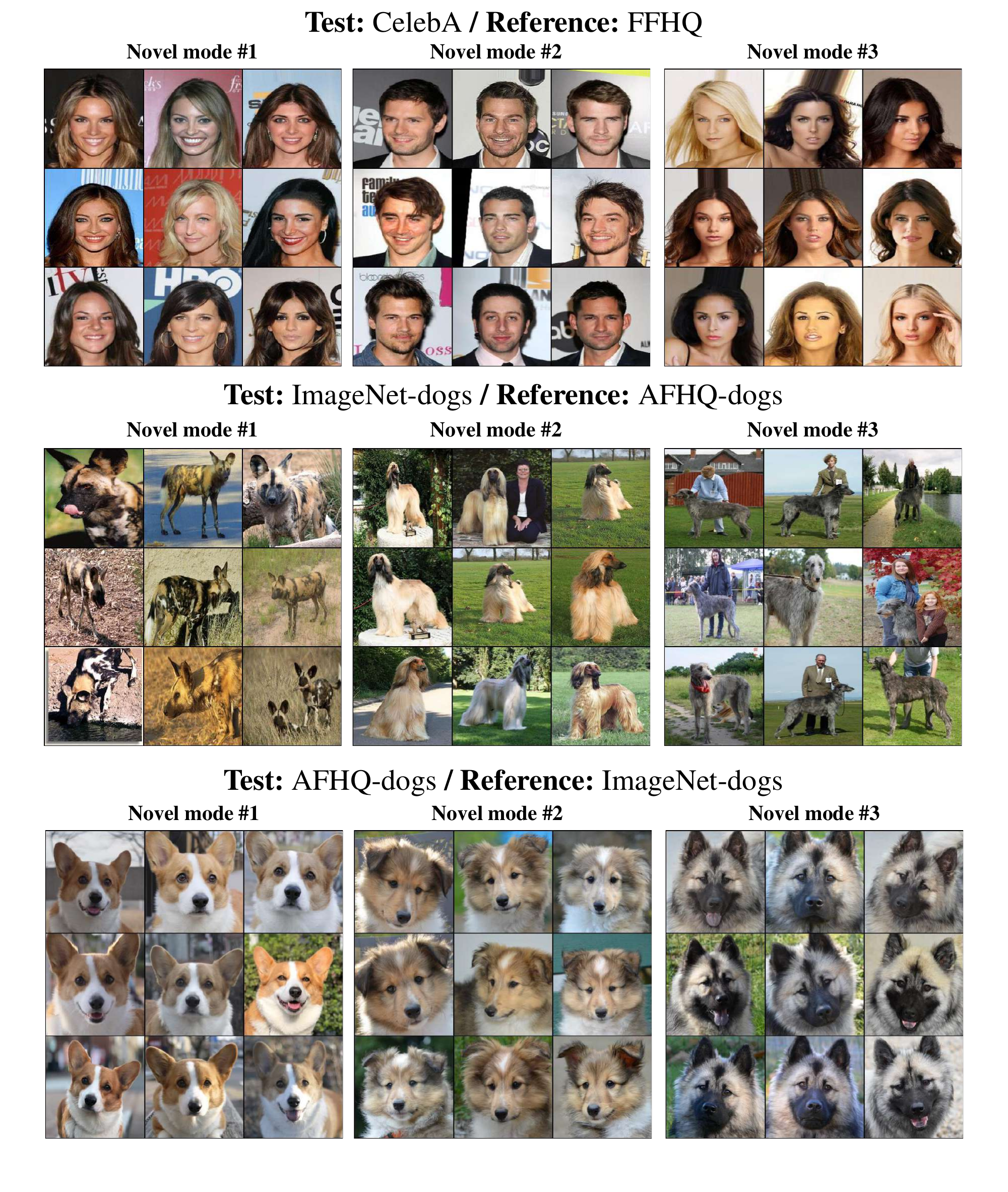}
    \caption{Novel modes between real datasets visualized with top-3-ranked eigenvectors. Extra samples of Figure \ref{fig:visual_realmode}. Inception-V3 embedding is used.}
    \label{fig:supp_real_novel_mode}
\end{figure*}

\begin{figure*}
    \centering
    \includegraphics[width=\textwidth]{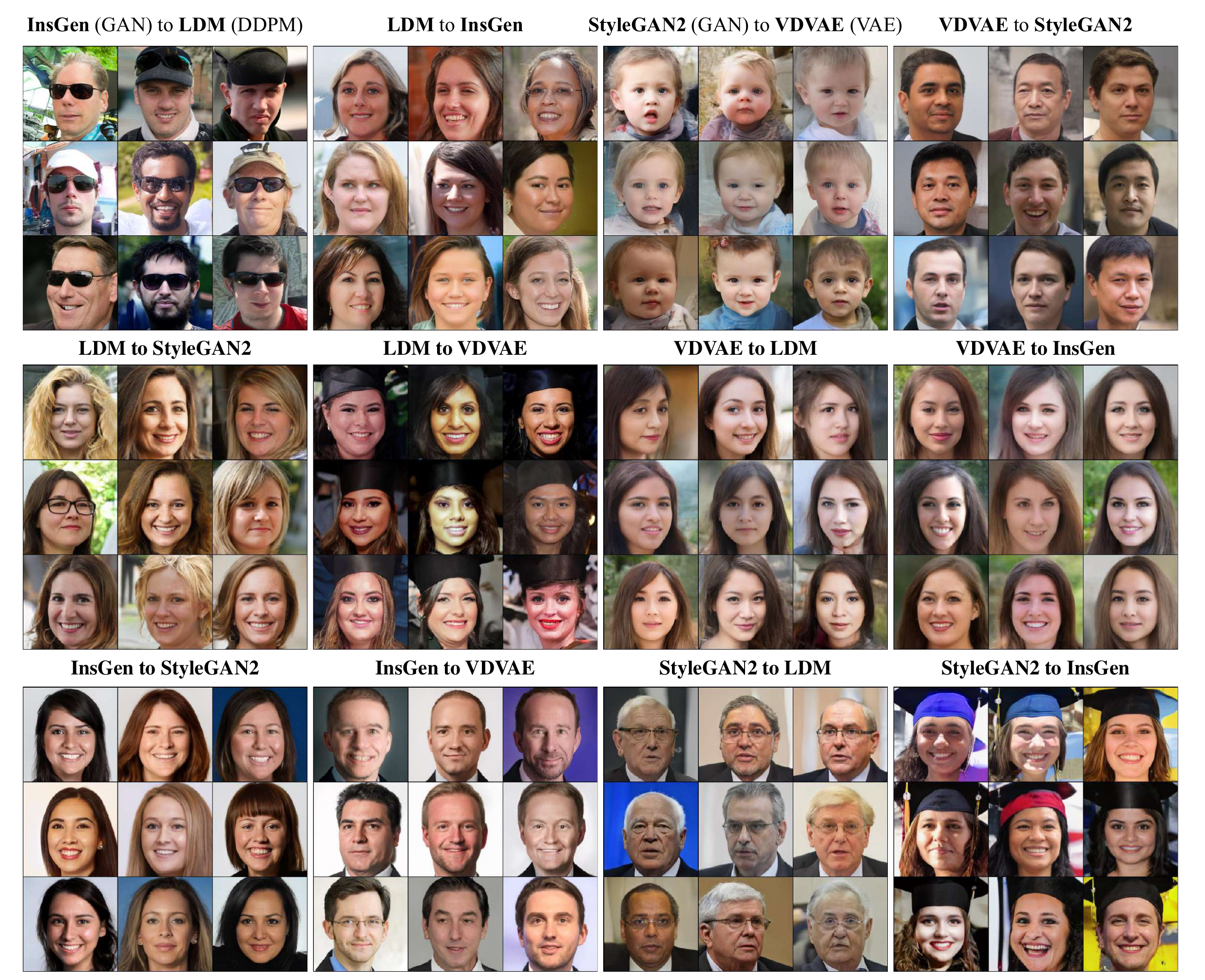}
    \caption{Novel modes between FFHQ-trained generative models in various architecture with the top-ranked eigenvector. Extra samples of Figure \ref{fig:visual_genmode}. Inception-V3 embedding is used.}
    \label{fig:supp_gen_novel_mode}
\end{figure*}

\textbf{Additional real dataset results.} Figure~\ref{fig:supp_real_novel_mode} shows detected novel modes between more real datasets. The novel modes of CelebA to FFHQ relate to the background of celebrities. For dog subsets of ImageNet and AFHQ, ImageNet-dogs seems to be novel in the dog breeds, while AFHQ-dogs seem to have more young dogs than ImageNet-dogs. Figure~\ref{fig:supp_gen_novel_mode} shows novel modes of all possible pairs of generative models in Figure~\ref{fig:visual_genmode}.

\begin{figure*}
    \centering
    \includegraphics[width=\textwidth]{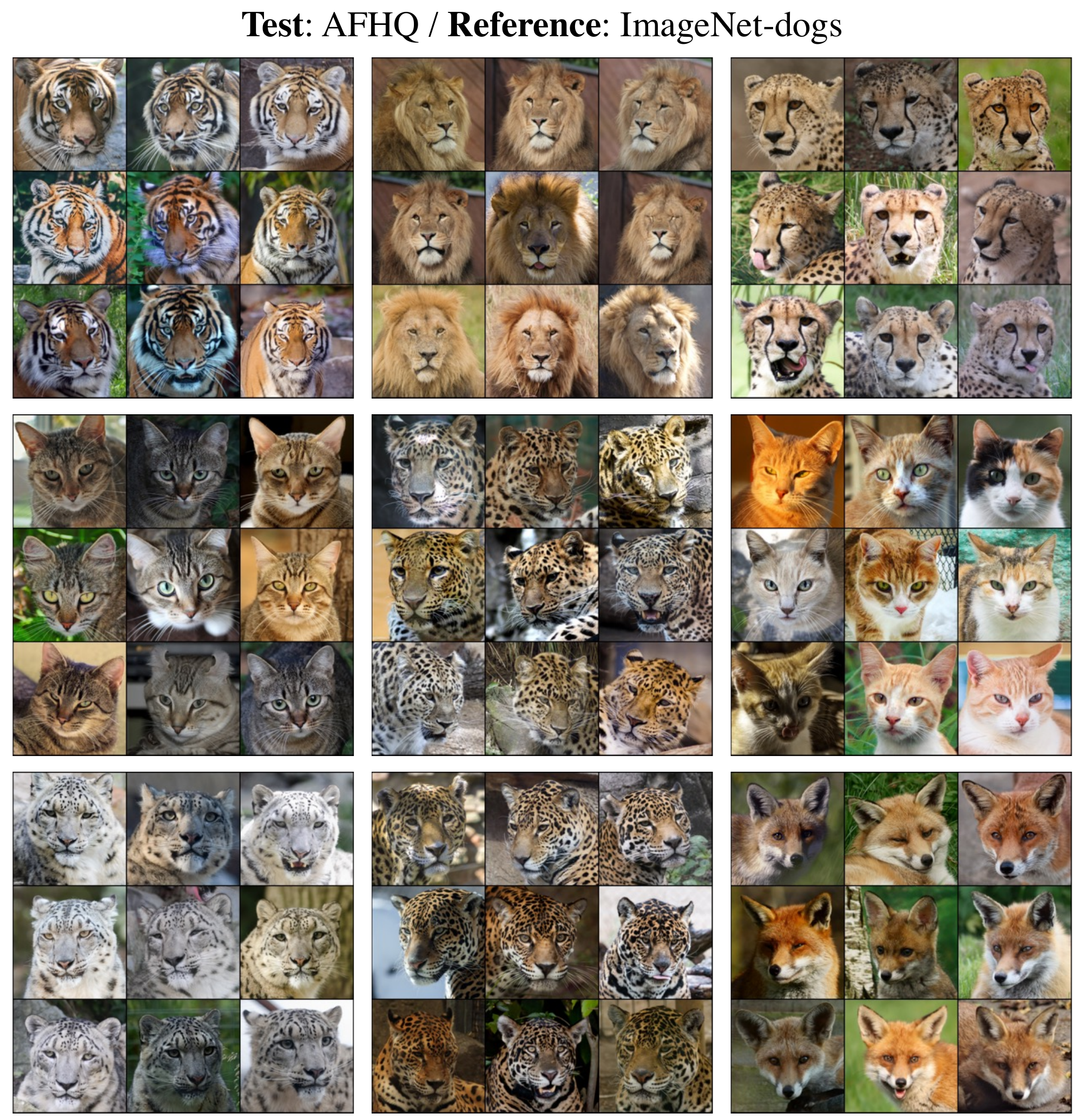}
    \caption{Top 9 novel modes of the AFHQ dataset w.r.t. the ImageNet-dogs dataset. Inception embedding is used.}
    \label{fig:supp_inception_afhq_dogs}
\end{figure*}

\begin{figure*}
    \centering
    \includegraphics[width=\textwidth]{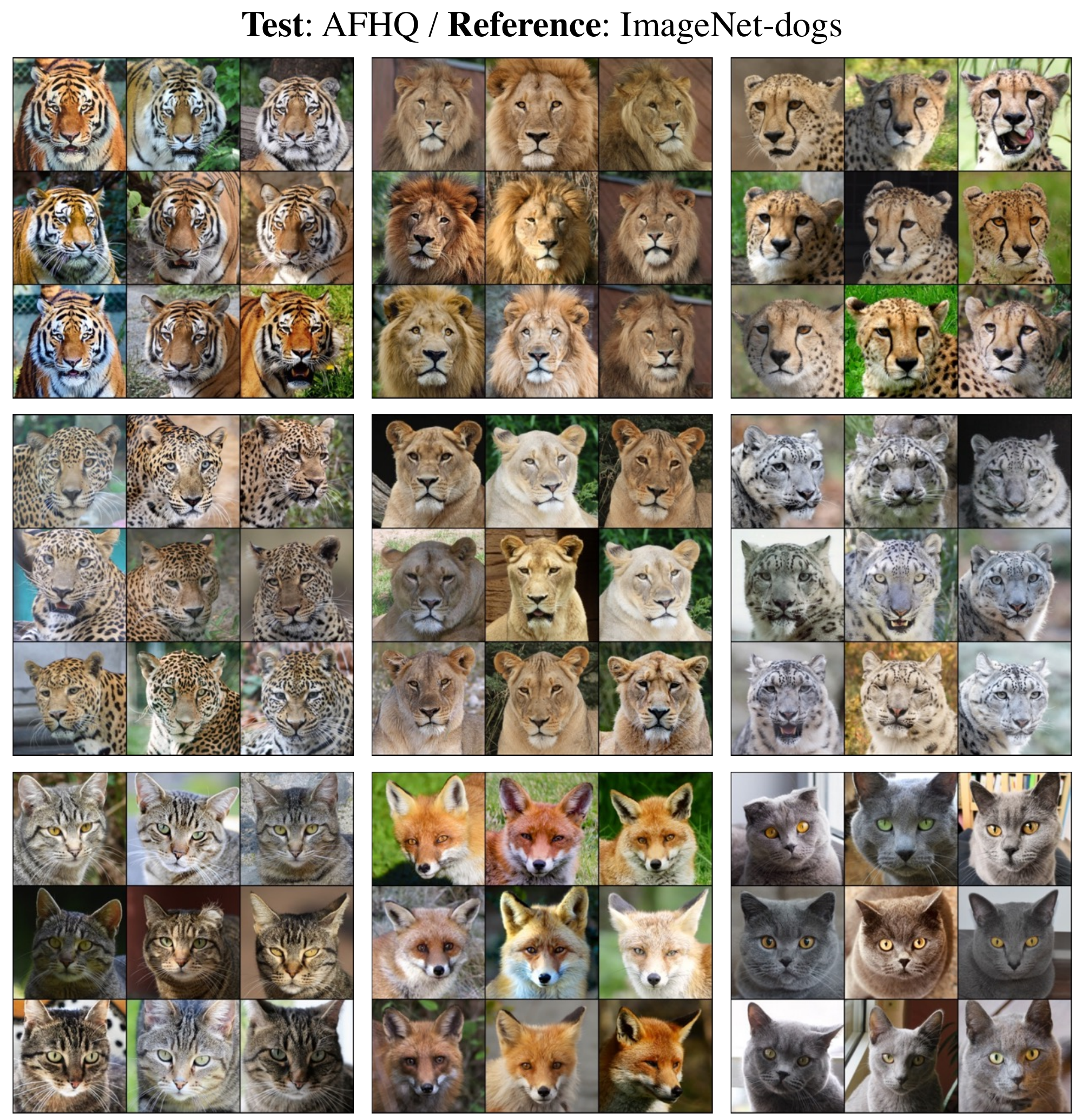}
    \caption{Top 9 novel modes of the AFHQ dataset w.r.t. the ImageNet-dogs dataset. DINOv2 embedding is used.}
    \label{fig:supp_dinov2_afhq_dogs}
\end{figure*}

\begin{figure*}
    \centering
    \includegraphics[width=\textwidth]{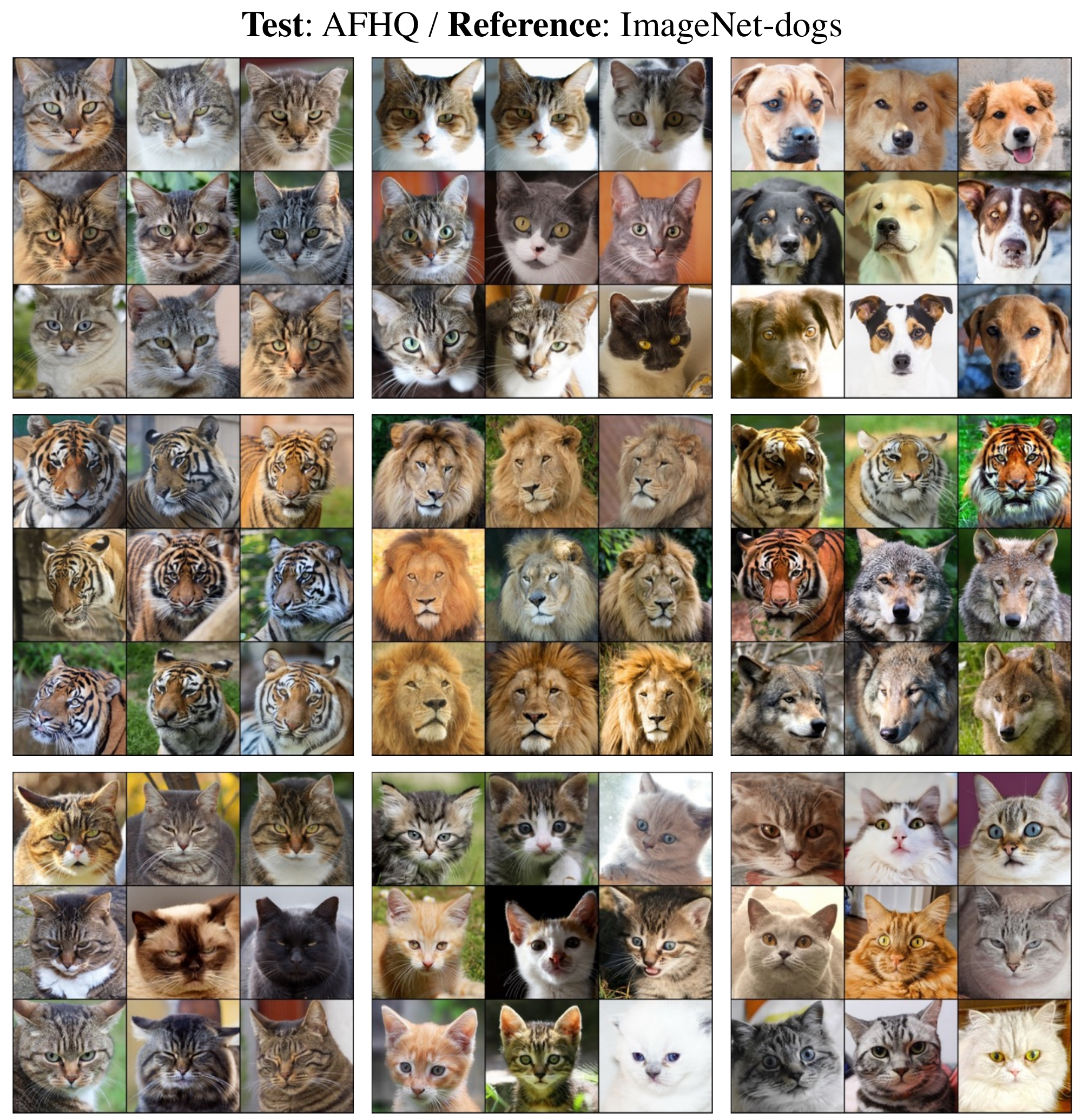}
    \caption{Top 9 novel modes of the AFHQ dataset w.r.t. the ImageNet-dogs dataset. CLIP embedding is used.}
    \label{fig:supp_clip_afhq_dogs}
\end{figure*}

\textbf{Novel modes detection with different embedding.} Figure \ref{fig:supp_inception_afhq_dogs}, \ref{fig:supp_dinov2_afhq_dogs}, \ref{fig:supp_clip_afhq_dogs} shows the detected top-9 novel modes of the AFHQ dataset with respect to the ImageNet-dogs dataset with Inception-V3, DINOv2, and CLIP embedding, respectively. The choice of embedding  affect the detected novel modes and their rankings by the proposed KEN method.

\textbf{Generative models' KEN scores with different embedding.} Table \ref{tab:benchmark_novelty_inception_supp}, \ref{tab:benchmark_novelty_dinov2_supp}, and \ref{tab:benchmark_novelty_clip_supp} shows KEN scores between generative models trained on the FFHQ dataset with Inception-V3, DINOv2, and CLIP embedding. We observed the rankings of the average KEN score are consistent with the same embedding but different choices of bandwidth parameter $\sigma$. However, the average KEN score ranking of generative models evaluated by the Inception-V3 embedding is different from rankings evaluated with DINOv2 and CLIP embedding.

\begin{table*}[t]
    \caption{Benchmarking fitness of generative models. More powerful models tend to have lower KEN to the training set. $\sigma=10$. Inception-V3 embedding is used.}
    \label{tab:benchmark_fitness}
    \centering
    \scalebox{0.95}{\begin{tabular}{clrrrrrrr}
        \toprule
        Dataset                    & Model           & IS     & FID   & Precision & Recall & Density & Coverage & KEN  \\
        \midrule
                                   & GGAN \cite{lim2017ggan}           & 6.51   & 40.22 & 0.56  & 0.30    & 0.42    & 0.31     & 3.03 \\
                                   & DCGAN \cite{radford2015dcgan}          & 5.76   & 51.98 & 0.62  & 0.16    & 0.56    & 0.25     & 3.02 \\
                                   & WGAN-WC \cite{arjovsky2017wgan-wc}        & 3.99   & 95.69 & 0.53  & 0.04    & 0.40     & 0.11     & 3.01 \\
                                   & WGAN-GP \cite{gulrajani2017wgan-gp}         & 7.04   & 26.42 & 0.62 & 0.56     & 0.55    & 0.46     & 2.99 \\
                                   & ACGAN \cite{odena2017acgan}          & 7.02   & 35.42 & 0.60  & 0.23     & 0.50     & 0.32     & 2.99 \\
                                   & LSGAN \cite{mao2017lsgan}          & 7.13   & 31.31 & 0.61  & 0.41    & 0.50     & 0.42     & 2.97 \\
                                   & LOGAN \cite{wu2019logan}          & 7.95   & 17.86 & 0.64  & 0.64    & 0.60     & 0.56     & 2.90 \\
                                   & SAGAN  \cite{zhang2019sagan}         & 8.67   & 9.58  & 0.69 & 0.63     & 0.72    & 0.72     & 2.70 \\
                                   & SNGAN  \cite{miyato2018sngan}         & 8.77   & 8.50   & 0.71 & 0.62     & 0.79    & 0.75     & 2.65 \\
                                   & BigGAN \cite{brock2018large}         & 9.14   & 6.80   & 0.71 & 0.61     & 0.86    & 0.80      & 2.59 \\
                                   & ContraGAN \cite{kang2021contragan}      & 9.40    & 6.55  & 0.73 & 0.61     & 0.87    & 0.81     & 2.57 \\
    \multirow{-12}{*}{CIFAR10}     & StyleGAN2-ADA \cite{karras2020stylegan2-ada}     & 10.14  & 3.61  & 0.73 & 0.67     & 0.98    & 0.89     & 2.50 \\
         \midrule
                                   & SAGAN \cite{zhang2019sagan}          & 14.47  & 64.04 & 0.33  & 0.54    & 0.16    & 0.14     & 3.46 \\
                                   & StyleGAN2-SPD \cite{karras2019style}   & 21.08  & 35.27 & 0.50 & 0.62      & 0.37    & 0.33     & 3.17 \\
                                   & StyleGAN3-t-SPD \cite{Karras2021stylegan3-t} & 20.90   & 33.69 & 0.52 & 0.61     & 0.38    & 0.32     & 3.13 \\
                                   & SNGAN \cite{miyato2018sngan}          & 32.28  & 28.66 & 0.54 & 0.67     & 0.42    & 0.41     & 3.07 \\
                                   & ContraGAN \cite{kang2021contragan}      & 25.19  & 28.33 & 0.67 & 0.53     & 0.64    & 0.34     & 2.91 \\
                                   & ReACGAN  \cite{kang2021reacgan}       & 52.95  & 18.19 & 0.76 & 0.40     & 0.88    & 0.49     & 2.67 \\
                                   & BigGAN-2048 \cite{brock2018large}    & 104.57 & 11.92 & 0.74 & 0.40     & 0.98    & 0.75     & 2.56 \\
    \multirow{-9}{*}{\makecell{ImageNet \\ $128^2$}} & StyleGAN-XL \cite{sauer2022stylegan}     & 225.16 & 2.71  & 0.80 & 0.63      & 1.12    & 0.93     & 2.42 \\
    \bottomrule
    \end{tabular}}
\end{table*}

\begin{table*}
    \caption{FFHQ-trained generative models' pairwise KEN score. Inception-V3 embedding is used.}
    \label{tab:benchmark_novelty_inception_supp}
    \centering
    \scalebox{0.9}{
    \begin{tabular}{clcccccr}
    \toprule
    \multirow{2}{*}{\makecell{Bandwidth \\ $\sigma$}} & \multirow{2}{*}{\makecell{Generative Models \\ (Test Models)}} & \multicolumn{5}{c}{Reference Models}                \\
    \cmidrule(){3-7} 
                                      & & InsGen  & StyleGAN-XL  & StyleGAN2 & LDM & VDVAE & Avg. KEN \\
    \midrule
    \multirow{5}{*}{10} & InsGen \cite{yang2021insgen}                            & -      & 3.56 & 3.55      & 3.64  & 4.27        & 3.76     \\
    & StyleGAN-XL  \cite{sauer2022stylegan}                               & 3.57   & -    & 3.61      & 3.61  & 4.21        & 3.75     \\
    & StyleGAN2 \cite{karras2019style}                        & 3.46   & 3.54 & -         & 3.60  & 4.08        & 3.67     \\
    & LDM \cite{rombach2021highresolution}                             & 3.45   & 3.45 & 3.49      & -     & 3.97        & 3.59     \\
    & VDVAE \cite{child2020very}                      & 3.26   & 3.24 & 3.19      & 3.17  & -           & 3.22      \\

    \midrule
    \multirow{5}{*}{15} & InsGen \cite{yang2021insgen}                            & -      & 1.17 & 1.18      & 1.26  & 1.87        & 1.37     \\
    & StyleGAN-XL  \cite{sauer2022stylegan}                               & 1.16   & -    & 1.19      & 1.24  & 1.83        & 1.36     \\
    & StyleGAN2 \cite{karras2019style}                        & 1.12   & 1.18 & -         & 1.26  & 1.76        & 1.33     \\
    & LDM \cite{rombach2021highresolution}                             & 1.09   & 1.08 & 1.14      & -     & 1.59        & 1.23     \\
    & VDVAE \cite{child2020very}                      & 0.96   & 0.95 & 0.94      & 0.91  & -           & 0.94      \\
    
    \bottomrule
    \end{tabular}}
\end{table*}

\begin{table*}
    \caption{FFHQ-trained generative models' pairwise KEN score. DINOv2 embedding is used.}
    \label{tab:benchmark_novelty_dinov2_supp}
    \centering
    \scalebox{0.9}{
    \begin{tabular}{clcccccr}
    \toprule
    \multirow{2}{*}{\makecell{Bandwidth \\ $\sigma$}} & \multirow{2}{*}{\makecell{Generative Models \\ (Test Models)}} & \multicolumn{5}{c}{Reference Models}                \\
    \cmidrule(){3-7} 
                                      & & StyleGAN-XL  & LDM  & InsGen & StyleGAN2 & VDVAE & Avg. KEN \\

    \midrule
    \multirow{5}{*}{30} & StyleGAN-XL  \cite{sauer2022stylegan}                           & -      & 4.35 & 4.53      & 4.59  & 4.75        & 4.56     \\
    & LDM \cite{rombach2021highresolution}                              & 4.25   & -    & 4.28      & 4.37  & 4.52        & 4.36     \\
    & InsGen \cite{yang2021insgen}                        & 4.39   & 4.24 & -         & 3.92  & 4.70        & 4.31     \\
    & StyleGAN2 \cite{karras2019style}                             & 4.38   & 4.27 & 3.87      & -     & 4.64        & 4.29     \\
    & VDVAE \cite{child2020very}                      & 4.17   & 4.02 & 4.22      & 4.24  & -           & 4.16      \\

    \midrule
    \multirow{5}{*}{50} & StyleGAN-XL  \cite{sauer2022stylegan}                           & -      & 1.48 & 1.60      & 1.65  & 1.84        & 1.64     \\
    & LDM \cite{rombach2021highresolution}                              & 1.41   & -    & 1.46      & 1.53  & 1.71        & 1.53     \\
    & InsGen \cite{yang2021insgen}                        & 1.50   & 1.44 & -         & 1.25  & 1.83        & 1.51     \\
    & StyleGAN2 \cite{karras2019style}                             & 1.50   & 1.47 & 1.21      & -     & 1.79        & 1.49     \\
    & VDVAE \cite{child2020very}                      & 1.42   & 1.34 & 1.47      & 1.49  & -           & 1.43      \\
    
    \bottomrule
    \end{tabular}}
\end{table*}

\begin{table*}
    \caption{FFHQ-trained generative models' pairwise KEN score. CLIP embedding is used.}
    \label{tab:benchmark_novelty_clip_supp}
    \centering
    \scalebox{0.9}{
    \begin{tabular}{clcccccr}
    \toprule
    \multirow{2}{*}{\makecell{Bandwidth \\ $\sigma$}} & \multirow{2}{*}{\makecell{Generative Models \\ (Test Models)}} & \multicolumn{5}{c}{Reference Models}                \\
    \cmidrule(){3-7} 
                                      & & StyleGAN-XL  & LDM  & InsGen & StyleGAN2 & VDVAE & Avg. KEN \\
    \midrule
    \multirow{5}{*}{5} & StyleGAN-XL  \cite{sauer2022stylegan}                           & -      & 4.11 & 3.82      & 3.92  & 4.26        & 4.03     \\
    & LDM \cite{rombach2021highresolution}                              & 4.03   & -    & 3.85      & 3.80  & 4.30        & 4.00     \\
    & InsGen \cite{yang2021insgen}                        & 3.80   & 3.88 & -         & 3.64  & 4.26        & 3.90     \\
    & StyleGAN2 \cite{karras2019style}                             & 3.84   & 3.77 & 3.57      & -     & 4.10        & 3.82     \\
    & VDVAE \cite{child2020very}                      & 3.49   & 3.58 & 3.49      & 3.46  & -           & 3.51      \\

    \midrule
    \multirow{5}{*}{10} & StyleGAN-XL  \cite{sauer2022stylegan}                           & -      & 0.90 & 0.77      & 0.82  & 1.10        & 0.90     \\
    & LDM \cite{rombach2021highresolution}                              & 0.87   & -    & 0.79      & 0.79  & 1.11        & 0.89     \\
    & InsGen \cite{yang2021insgen}                        & 0.79   & 0.82 & -         & 0.72  & 1.12        & 0.86     \\
    & StyleGAN2 \cite{karras2019style}                             & 0.80   & 0.78 & 0.69      & -     & 1.05        & 0.83     \\
    & VDVAE \cite{child2020very}                      & 0.73   & 0.75 & 0.73      & 0.73  & -           & 0.74      \\
    
    \bottomrule
    \end{tabular}}
\end{table*}

\end{document}